\documentclass[sigplan,nonacm,screen]{acmart}
\author{Yuyang Huang}
\authornote{Part of work is done when intern at Microsoft Research}
\affiliation{%
  \institution{The University of Chicago\\Microsoft Research}
  \city{Chicago, IL}
  \country{USA}
}

\author{Yuhan Liu}
\affiliation{%
  \institution{The University of Chicago}
  \city{Chicago, IL}
  \country{USA}
}

\author{Haryadi S. Gunawi}
\affiliation{%
  \institution{The University of Chicago}
  \city{Chicago, IL}
  \country{USA}
}

\author{Beibin Li}
\authornote{Moved to Meta}
\affiliation{%
  \institution{Microsoft Research}
  \city{Redmond, WA}
  \country{USA}
}

\author{Changho Hwang}
\affiliation{%
  \institution{Microsoft Research}
  \city{Vancouver, BC}
  \country{Canada}
}

\renewcommand{\shortauthors}{Huang et al.}

\usepackage{relsize}
\usepackage{enumerate}
\usepackage{graphicx}
\usepackage{url}  
\usepackage{multirow}
\usepackage{arydshln}

\hypersetup{colorlinks=true,
  citecolor=Maroon,
  linkcolor=Green,
  urlcolor=Maroon}

\usepackage{rotating}
\usepackage{xspace}

\usepackage{amsmath}  
\usepackage{mathtools}

\usepackage{floatflt}
\usepackage{wrapfig}
\usepackage{alltt}
\usepackage{epstopdf}
\usepackage{subfigure}

\usepackage{physics}

\usepackage{fancyvrb}
\usepackage[normalem]{ulem} 

\usepackage{etoolbox}
\apptocmd{\thebibliography}{\raggedright}{}{}

\usepackage{colortbl} 

\usepackage{wasysym}  
\usepackage{pifont}   

\usepackage{upgreek}

\usepackage{tikz, pgfplots, pgfplotstable}
\usetikzlibrary{arrows.meta, positioning, calc}
\usepgfplotslibrary{fillbetween}
\usepackage{upgreek}

\begin{document}

\def \us {$\upmu$s}  
\def \yes {$\surd$}  

\def \none {N$_1$}
\def \ntwo {N$_1$}
\def \ntri {N$_1$}
\def \nnnn {N$_n$}

\def \ra {$\rightarrow$}

\def \tms {$\times$}

\def \whitecircle {$\ocircle$}
\def \blackcircle {\ding{108}}
\def \whitesquare {$\Box$}
\def \blacksquare {\ding{110}}
\def \whitediamond {$\Diamond$}
\def \blackdiamond {\ding{117}}
\def \whitetriangle {$\bigtriangleup$}
\def \blacktriangle {\ding{115}}
\def \whitedtriangle {$\bigtriangledown$}
\def \blackdtriangle {\ding{116}}
\def \rtb {\ding{253}}

\newcommand{\tc}[1]{\textbf{\textcircled{#1}}}
\def \tcone {\ding{202}\xspace}
\def \tctwo {\ding{203}\xspace}
\def \tctri {\ding{204}\xspace}
\def \tcfour {\ding{205}\xspace}
\def \tcfive {\ding{206}\xspace}

\def \cm {\checkmark}


\newcommand{\yarn}[1]{\jira{YARN}{yarn}{#1}}
\newcommand{\ca}[1]{\jira{CASSANDRA}{ca}{#1}}
\newcommand{\hb}[1]{\jira{HBASE}{hb}{#1}}
\newcommand{\mr}[1]{\jira{MAPREDUCE}{mr}{#1}}
\newcommand{\zk}[1]{\jira{ZOOKEEPER}{zk}{#1}}
\newcommand{\ha}[1]{\jira{HADOOP}{hadoop}{#1}}

\newcommand{\jira}[3]{\href{http://issues.apache.org/jira/browse/#1-#3}{#2#3}}

\newcommand{\vtwenty}{\vspace{20pt}}
\newcommand{\vfifteen}{\vspace{15pt}}
\newcommand{\vten}{\vspace{10pt}}
\newcommand{\vfive}{\vspace{5pt}}
\newcommand{\vthree}{\vspace{3pt}}

\newcommand{\vminfive}{\vspace{-5pt}}
\newcommand{\vminten}{\vspace{-10pt}}
\newcommand{\vminfifteen}{\vspace{-15pt}}
\newcommand{\vmintwenty}{\vspace{-20pt}}

\def \hmina {\hspace{-0.1in}}
\def \hminb {\hspace{-0.2in}}

\newcommand{\ub}[1]{\underline{{\bf #1}}}
\newcommand{\ts}[1]{{\tt{\small#1}}}
\newcommand{\bquote}{\vspace{-0.25cm} \begin{quote}}
    \newcommand{\equote}{\end{quote}\vspace{-0.2cm} }
\def \sec {\S}
\def \yes {$\surd$}

\def \nospace {
  \setlength{\itemsep}{0pt}
  \setlength{\parskip}{0pt}
  \setlength{\parsep}{0pt}
}

\newcommand{\myquote}[1]{
  \begin{quote}
    \centering
    \small
    \textit{#1}
  \end{quote}
}

\newenvironment{enumerate2}{
  \begin{enumerate} \vminfive
    \setlength{\itemsep}{1pt}
    \setlength{\parskip}{0pt}
    \setlength{\parsep}{0pt}
    }{
  \end{enumerate}
}

\newenvironment{itemize2}{
  \begin{itemize} \vminfive
    \renewcommand{\labelitemi}{-}
    \setlength{\itemsep}{1pt}
    \setlength{\parskip}{0pt}
    \setlength{\parsep}{0pt}
    }{
  \end{itemize}
}

\newenvironment{packeditemize}{\begin{list}{$\bullet$}{\setlength{\itemsep}{0.5pt}\addtolength{\labelwidth}{-4pt}\setlength{\leftmargin}{2ex}\setlength{\listparindent}{\parindent}\setlength{\parsep}{1pt}\setlength{\topsep}{2pt}}}{\end{list}}


\newcommand{\notes}[1]{\textcolor{darkgray}{{\footnotesize {\em (Notes: #1)}}}}

\newcommand{\student}[1]{\textcolor{purple}{{\footnotesize {\bf (STU: #1)}}}}

\newcommand{\border}[1]{\textbf{\textcolor{Cyan}{[BDR]#1}}}
\newcommand{\red}[1]{\textcolor{red}{#1}}

\newcounter{hsgcounter}
\setcounter{hsgcounter}{1}
\newcommand{\hsg}[1]{{\footnotesize
      \textbf{\textcolor{red}{(HSG$_{\arabic{hsgcounter}}$: #1)}}}
  \stepcounter{hsgcounter}}

\newcounter{chhcounter}
\setcounter{chhcounter}{1}
\newcommand{\chh}[1]{{\footnotesize
      \textbf{\textcolor{red}{(CHH$_{\arabic{chhcounter}}$: #1)}}}
  \stepcounter{chhcounter}}

\newcounter{bblcounter}
\setcounter{bblcounter}{1}
\newcommand{\bbl}[1]{{\footnotesize
      \textbf{\textcolor{purple!70}{(BBL$_{\arabic{bblcounter}}$: #1)}}}
  \stepcounter{bblcounter}}

\newcounter{roycounter}
\setcounter{roycounter}{1}
\newcommand{\roy}[1]{{\footnotesize
      \textbf{\textcolor{blue}{(ROY$_{\arabic{roycounter}}$: #1)}}}
  \stepcounter{roycounter}}

\newcommand{\pc}[1]{\textcolor{blue}{\textit{(PC: #1)}}} 
\newcommand{\todo}[1]{\textcolor{red}{{\footnotesize {\bf (TODO: #1)}}}}

\newcommand{\newtxt}[1]{\textcolor{CornflowerBlue}{#1}} 
\newcommand{\oldtxt}[1]{\textcolor{gray}{{\footnotesize {\em OLD TEXT: #1}}}}
\newcommand{\bluetxt}[1]{\textcolor{blue}{#1}}
\newcommand{\rbt}[1]{\textcolor{red}{\textbf{#1}}}
\newcommand{\bbt}[1]{\textcolor{blue}{\textbf{#1}}}




\def \vvvnb {\vfifteen \noindent $\bullet$~}
\def \vvnb {\vten \noindent $\bullet$~}
\def \vnb {\vfive \noindent $\bullet$~}
\def \vn {\vfive \noindent}

\def \mb {\vspace{8pt}\nb}
\def \tb {\vspace{8pt}\nb}

\def \vvni {\vten \noindent}
\def \vni {\vfive \noindent}
\def \nb {\noindent $\bullet$~}
\def \ni {\noindent}
\def \bb {$\bullet$~}

\newcommand{\hypo}[1]{
  \begin{quote}
    \stepcounter{HYPO}{\bf Hypothesis \arabic{HYPO}:}
    {\em #1}
  \end{quote}
}

\newcommand{\taskformat}[2]{#1\textsc{#2}}

\newcommand{\task}[3]{
  \begin{quote}
    \phantomsection
    \hypertarget{task#1#2}{}
    {\bf Task \taskformat{#1}{#2}:}
    {\em #3}
  \end{quote}
}

\newcommand{\tasklink}[2]{\hyperlink{task#1#2}{\taskformat{#1}{#2}}}

\newcounter{HYPO}
\newcounter{TASK}

\newcommand{\rs}{{ResearchStaff$_1$}}
\newcommand{\pd}{{\bf Postdoc$_1$}}
\newcommand{\raOne}{{\bf RA$_1$}}
\newcommand{\raTwo}{{\bf RA$_2$}}
\newcommand{\ndv}{{\bf NDV}}
\newcommand{\ug}{{\bf Undergrad$_1$}}


\newcommand{\sssubsection}[1]{\vten\ni\textbf{\large{\textsc{#1}}}}

\newcounter{mysubcounter}
\setcounter{mysubcounter}{1}
\newcommand{\mysub}[1]{\vfive \noindent \textsc{\textbf{#1:}}}

\newcommand{\emptypage}{
  \newpage
  (empty page)
}

\newcommand{\myrotate}[1]{\begin{rotate}{90} {\bf #1} \end{rotate}}

\newcommand{\mycaption}[3]{
  \caption{
    \label{#1}
    {\bf #2. }
    {\em \small #3}
  }
}

\newcommand{\eg}{\textit{e.g.}}
\newcommand{\ie}{\textit{i.e.}}
\newcommand{\etal}{\textit{et al.}}
\newcommand{\etc}{etc.}

\newcommand{\sstar}{$^{*}$}
\newcommand{\stwostars}{$^{**}$}
\newcommand{\stristars}{$^{***}$}
\newcommand{\srealstar}{$^{\star}$}
\newcommand{\sdag}{$^{\dag}$}
\newcommand{\sddag}{$^{\ddag}$}

\newcommand{\tool}{\textsc{Alchemist}\xspace}

\newcounter{Xcounter}
\setcounter{Xcounter}{1}
\newcommand{\xxxreset}{\setcounter{Xcounter}{1}}
\newcommand{\xxx}{{\footnotesize
      \textcolor{red}{
        \textbf{xxx$_{\arabic{Xcounter}}$}\stepcounter{Xcounter}}~}}

\newcommand{\xxxinfig}{\textcolor{red}{\textbf{xx}}} 

\newcounter{Qcounter}
\setcounter{Qcounter}{1}
\newcommand{\qqq}{{\footnotesize
      \textcolor{red}{
        \textbf{qqq$_{\arabic{Qcounter}}$}\stepcounter{Qcounter}}~}}


\newcounter{Fcounter}
\newcommand{\freset}{\setcounter{Fcounter}{1}}

\newcommand{\finding}[1]{
  \begin{spacing}{0.80}
    \findingTable{#1}
  \end{spacing}
}

\definecolor{fgray}{gray}{0.9}

\newcommand{\findingTable}[1]{
  \begin{table}[h!]
    \begin{tabular}{|p{3.2in}|}
      \hline
      \rowcolor{fgray}
      \findingBody{#1} \\
      \hline
    \end{tabular}
  \end{table}
  \vminten
}

\newcommand{\findingBody}[1]{{\small
      \textbf{Finding \#$\arabic{Fcounter}$:}
      \stepcounter{Fcounter}
      #1 }
}

\setcounter{Fcounter}{1}


\newcommand{\summtitle}{
  \setcounter{page}{0}
  \thispagestyle{empty}
  \ni {\bf\Large Proposal Title: \drsys: Drill-Ready Cloud Computing}
  \vfifteen
}

\newcommand{\summintro}{
  \setcounter{page}{0}
  \thispagestyle{empty}
  \ni {\bf\Large Project Summary}
  \vfive
}

\newcommand{\supsection}[1]{\noindent{\Large{\bf #1}}\vten}

\newcommand{\summaryheader}{
  \thispagestyle{empty}
  \noindent {\bf\Large B $\;$ Project Summary}
}

\newcommand{\bodyheader}{
  \setcounter{page}{1}
  \noindent {\bf\Large D $\;$ Project Description}
}

\newcommand{\myparashort}[1]{\vspace{0.05cm}\noindent{\bf {#1}}~}
\newcommand{\mypara}[1]{\vspace{0.05cm}\noindent{\bf {#1}:}~}
\newcommand{\myparatight}[1]{\vspace{0.02cm}\noindent{\bf {#1}:}~}
\newcommand{\myparaq}[1]{\smallskip\noindent{\bf {#1}?}~}
\newcommand{\myparaittight}[1]{\smallskip\noindent{\emph {#1}:}~}
\newcommand{\myparaqtight}[1]{\smallskip\noindent{\bf {#1}}~}


\newcommand{\sysname}{\textsc{Alchemist}\xspace}

\date{}

\title{\sysname: Towards the Design of Efficient Online Continual Learning System}

\def \myabstract{Abstract here.}

\begin{abstract}
Continual learning has become a promising solution to refine large language models incrementally by leveraging user feedback.
In particular, online continual learning --- iteratively training the model with small batches of user feedback --- has demonstrated notable performance improvements. 
However, the existing practice of separating training and serving processes forces the online trainer to recompute the intermediate results already done during serving. Such redundant computations can account for 30\%–42\% of total training time.

In this paper, we propose \sysname, to the best of our knowledge, the first online continual learning system that efficiently reuses serving activations to increase training throughput.
\sysname introduces two key techniques: (1) recording and storing activations and KV cache only during the prefill phase to minimize latency and memory overhead; and (2) smart activation offloading and hedging.
Evaluations with inputs of varied token length sampled from ShareGPT dataset show that compared with a separate training cluster, \sysname significantly increases training throughput by up to 1.72x, reduces up to 47\% memory usage during training, and supports up to 2x more training tokens --- all while maintaining negligible impact on serving latency.

\end{abstract}
\vten


\maketitle

\section{Introduction}

Large Language Model (LLM) services, widely adopted in modern cloud ecosystems, power applications, such as chatbots, search engines, recommendation systems, and API offerings. The market is projected to grow from \$6.5B (2024) to \$140.8B (2033), as 92\% of Fortune 500 companies integrate cloud-hosted LLMs for generative AI workflows \cite{techtarget2023llm, exploding2024llm, mckinsey2023ai}.
To ensure these models remain accurate and continue to enhance user experiences, organizations must regularly retrain them to learn the latest data and refine performance.


One of the most popular and promising training approaches is continual learning~\cite{Gabriel_2020, copf.corr23, ernie.aaai20, reformulating.corr23, adaptinglargelanguagemodels.corr24, Gabriel_2020, cert.corr22, conpet.corr23, continualpretrainingmitigatesforgetting.corr22, dontstoppretrainingadapt.corr20}, due to its capability of allowing models to be refined frequently, without training from scratch. It incrementally adapts the models with up-to-date real-world information and interactions with users or models that mimic user preferences~\cite{onlineaifeedback.corr24, rlaif.corr24, deepseek, metacot}.
For example, chatbots with web searching capability can learn from the newest searched context (\eg, latest news, financial data, \etal) to provide the latest information in future responses. Or, code completion applications allow users to accept or reject suggested code, and this user feedback can be used to allow models to continuously learn their preference, providing better and more personalized code completion and debug suggestions.

One of the most popular continual learning approach, \emph{online continual learning} (\S \ref{motiv-online-cont-learn}), {\em iteratively trains} the model based on a small batch of user feedback or model-generated labels collected {\em each time} the model updated~\cite{onlineaifeedback.corr24, ofsdpo.corr24, learnfromuseredit.nips24, showdonttellaligning.corr24, thingscringeothersiterative.corr24, glam.icml23, sacglam.corr24}.
 

\begin{figure}[t]
  \centering
  \includegraphics[width=\linewidth]{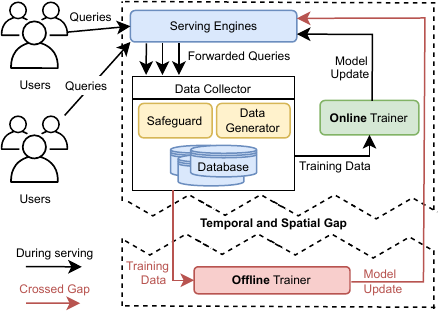}
  \vminten
  \mycaption{fig:overview}{Lifecycle of modern AI services}{}
  \vminten
\end{figure}
In this context, we observe a significant gap between the modern need for iterative and frequent updates in LLM services and the conventional paradigm for LLM training.
As illustrated in Figure~\ref{fig:overview}, existing systems typically separate training from serving to maintain low latency for serving.
This separation occurs both temporally (\ie, data collected during serving is only used for training after a considerable delay) and spatially (\ie, training is carried out on machines that are distinct from those used for serving).
However, such hard separation loses substantial opportunities to reduce computational costs.

An important and interesting observation we made is that, with such design, online continual learning trainer needs to {\em repeat the same computation that has already been conducted} by the serving process, and this redundant computation often contributes to a significant portion of the entire computation for training. 
In reality, the trainer would spend 30\%--42\% of the total time for the computations that have already been done during serving, depending on the loss functions.

This observation reveals a key insight of our approach: \emph{the intermediate results (\ie, activations) generated during serving can be directly re-used in the forward pass of training}. Specifically, we reuse these \emph{activations} from serving to eliminate redundant forward-pass computations.


However, if the training and serving processes are separate on different machines as existing systems designed, to reuse the serving actions, they need to be sent to the training machines via the network, 
posing large network transmission overhead given the large size of activations. 

An intuitive and straightforward design is to {\em remove} the hard separation between serving and training and co-locate serving and training on the same machine, leveraging GPU multiplexing techniques (\eg, either temporal sharing or spatial sharing).
So, once there is idle time in the serving process (\eg, during off-peak hours at night time), the serving activations can be used directly by the trainer, \emph{without} any transmission delay across machines.

A potential concern around this design is whether the serving workload offers sufficient idle resources. Nowadays, the serving resource is typically over-provisioned ~\cite{k8sautoscale} to ensure end users' experience (\eg latency service-level objectives (SLOs) guarantee). This indicates that a significant amount of idle resources is always available and can be harvested for training usage. 
Another concern is whether the newly collected serving data is safe to be immediately incorporated into model updates and whether the updated model can be immediately deployed to users. To address potential safety risks, multiple studies \cite{adversarialdpo.corr24, llamaguard.corr23, filtereddpo.corr24, probingtoxiccontent.acl21, beavertails.nips24} have proposed safeguards to ensure only safe data can be trained by the model. Many other works \cite{onlineaifeedback.corr24, ofsdpo.corr24, learnfromuseredit.nips24, showdonttellaligning.corr24, thingscringeothersiterative.corr24, glam.icml23, sacglam.corr24, reversekl.iclr23} focus on ensuring that the updated models neither drift too far away from the pre-trained model, losing its generality, nor generate unethical or unlawful responses.

However, there is still {\em a system-level challenge} that remains unsolved. 
Specifically, co-locating serving and training on the same machine may incur extra overhead in saving and reusing the serving activations.
First of all, although reusing activations can reduce training latency, the serving process needs to record and store the activations produced by \emph{each layer} and the forward computation graph, to be used in the backward pass for gradient computation (\S\ref{design-cache}).
This procedure incurs additional latency overhead to the forward passes during serving, which could potentially violate serving latency SLO.
Concretely, such latency overhead can be up to 35\% when storing and recording activations and computation graphs.
Secondly, in a normal serving procedure, each layer's activations are overwritten by the next layers' activations to reduce overall memory footprint~\cite{vllm}. Hence, storing each layer's activations for reusing in training also incurs extra memory overhead and severely limits the serving capacity due to the large size of activations.

To address the above challenges, we propose \sysname, to the best of our knowledge, the \emph{first efficient system that reuses serving activations in training} to increase online training throughput, with minimal impact on both serving latency and capacity.
\sysname entails two techniques:
\begin{packeditemize}
    \item \textbf{Minimal activation recording during serving}: To minimize the activations recording overhead during LLM serving, we only record the activations when processing user input, \ie, prefill phase, and disable it when generating new tokens, \ie, decoding phase.
    This can be effective because in real-world scenarios~\cite{sharegpt}, the {\em total} decoding time is often the majority of the end-to-end time.
    Furthermore, in the case where activations are required for both input and model response, \sysname reuses the KV cache (one kind of intermediate results) generated by prefill in serving to avoid activations recording for the entire generation procedures while still managing to reuse part of the activations calculated during serving (\S\ref{design-cache}).
    \item \textbf{Offloading of serving activations}: To guarantee serving capacity, when the serving requires more GPU memory based on an offline memory capacity profiler, \sysname frees the serving activations in a {\em as needed} fashion.
    Based on the insight that the activations of earlier layers are needed at last during the backward pass, \sysname frees activations in {\em forward order} and loads them in {\em the reverse order of the layers}, maximizing the overlap between the backward computation and activations loading.
\end{packeditemize}

We compare \sysname with the baseline that separates the training and serving processes on different machines, and show that on input with varied token length sampled from a popular dataset, ShareGPT~\cite{sharegpt}:

\begin{packeditemize}
    \item \sysname increases the training throughput by 1.26x--1.72x, across two different example continual learning methods with three types of LLMs.
    \item Compared to na\"ive loss calculation, by removing duplication in memory with KV cache reuse, \sysname saves memory by 32\% -- 47\% depending on the trained token length and supports up to 2x more maximum trainable tokens.
    \item \sysname achieves the performance gains with very minimal impact on serving latency.
\end{packeditemize}

\section{Background \& Motivation}  
\label{motiv-intro}

\subsection{Model Inference and Tuning}
\label{motiv-model-inference}
Modern LLMs and Transformers rely on transformer architectures~\cite{transformers}, where inference occurs in two phases.
During the \textbf{prefill phase}, the model processes the entire input prompt (e.g., a user’s query) in parallel, computing \textit{key-value (KV) caches} that capture contextual relationships between tokens. These caches persist to accelerate subsequent steps.  
During the \textbf{decoding phase}, the model generates output tokens auto-regressively by using the KV caches. Each new token’s attention scores reuse the cached keys/values from the prefill, minimizing redundant computation.  

In both phases, the model computes \textbf{activations}, the intermediate outputs from each layer’s linear operations (e.g., matrix multiplications) and non-linearities (e.g., ReLU). As illustrated in Figure \ref{fig-forward-backward}, these activations are critical for training: back-propagation uses them to compute gradients. However, serving systems discard activations immediately after inference to save memory, wasting computational work that training could reuse.

\begin{figure}[t]
  \centering
  \hspace{15pt}
  \includegraphics[]{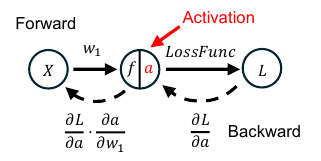}
  \mycaption{fig-forward-backward}{Activations in forward and backward passes of model training}{Activations, $a$, computed during forward pass will be stored and reference during backward pass to avoid redundant recomputation. The figure does not fully reflect what actually happen during training but is simplified only for illustration purpose. \S\ref{motiv-model-inference}}
\end{figure}

\subsection{Parameter Efficient Fine-Tuning (PEFT)}
With the rapid growth in LLMs' size, the cost of continually adapting models with the new information also skyrockets due to the requirements of a large number of GPUs. \textbf{PEFT} methods~\cite{prefixtunning.acl21, prompttunning.emnlp21, vera.iclr24, oft.nips23, ptuning.corr23, cpt.corr24, multitaskprompttuning.corr23} mitigate training costs by freezing the base model and only updating low-rank adapters with much smaller parameter sizes, hence reducing the memory and computation requirements. Consequently, these methods, especially \textbf{LoRA}~\cite{lora.corr21}, have become the most favorable methods for fine-tuning LLMs.

\subsection{Continuous Adaptation}  %
\label{motiv-cont-adapt}
With the help of PEFT method like LoRA, LLM services increasingly adopt \textbf{continual learning} to keep updated with latest information. Two paradigms dominate:  

\mypara{Continual Pretraining}
Models ingest fresh data (news, user prompts, code commits) to update factual knowledge~\cite{ernie.aaai20, reformulating.corr23, adaptinglargelanguagemodels.corr24, Gabriel_2020, cert.corr22, continualpretrainingmitigatesforgetting.corr22, dontstoppretrainingadapt.corr20}. For example, a financial LLM might retrain daily on earnings reports to improve stock analysis.  

\mypara{Continual Preference Alignment}
User feedback (e.g., accepting/rejecting code suggestions) fine-tunes models to individual or collective preferences~\cite{copf.corr23, deepseek, generallanguageassistantlaboratory.corr21, ppo.corr17, finetuninglanguagemodelshuman.corr20, learningsummarizehumanfeedback.corr22, alignmentlanguageagents.corr21, aligninglanguagemodelspreferences.corr23} Techniques like Direct Preference Optimization (DPO)~\cite{dpo.corr24} excel here due to its simplicity --- it leverages pairwise preferences to align outputs {\em without} requiring to design and train a separate reward model.

\subsection{Online Continuous Learning}%
\label{motiv-online-cont-learn}

Typical procedure in continuous learning often aggregates data from the serving side over weeks or months, known as offline learning.
But these delayed updates face challenges, for instance: {\em concept drift} occurs as user preferences shift, such as changes in coding style trends; {\em factual obsolescence} arises when news, APIs, or regulations become outdated.

Ideally, we want the model to be continuously trained on almost a real-time
stream of information and feedback.
This means that, instead of infrequent, large updates, model is frequently retrained and updated with small batches of latest.
Users (or LLMs that mimic users' preferences), in turn, provide the latest feedback or information based on the response generated by the updated version to further update and improve the model's performance {\em iteratively}.

This cyclic and iterative training process is often referred to as \textbf{online continuous learning}~\cite{onlineaifeedback.corr24, ofsdpo.corr24, learnfromuseredit.nips24, showdonttellaligning.corr24, thingscringeothersiterative.corr24, glam.icml23, sacglam.corr24}.
Unlike the traditional offline continuous learning procedure which trains the models based on a single large batch of stationary data for every iteration, online continuous learning pipeline emphasizes its nature in {\em iterative improvement over time} based on a micro-batch of newest information and feedback collected each time after the model is updated.

\subsection{Online Continuous Learning System}%
\label{motiv-online-cont-learn-sys}

Implementing an online continuous learning system introduces the practical
question of {\em where} to run training. A straightforward solution might be
to maintain a separate training cluster, but this approach entails substantial
business and operational tradeoffs. A reserved training cluster can be
underutilized, tying up resources that could have been allocated to improve
serving capacity or quality. On the other hand, an on-demand training cluster
(spun up only when user-serving traffic is low) incurs orchestration
complexity and repeated overhead each time training is triggered. Both
approaches can lead to inefficiencies and added costs.

\subsubsection{Reusing Activations}
Furthermore, in an online continuous learning scenario, since user prompts or user feedback will be the training input (or part of the input), the forward passes during serving have already calculated the activations we need for backpropagation. Such forward passes can cost as much as 43\% of the total training time, shown in red color in Fig \ref{fig-train-breakdown}. 
Hence, intuitively, it is {\em more efficient to reuse} the activations calculated from the serving forward pass during backpropagation rather than re-calculate them again.

\begin{figure}[t]
  \centering
  \hspace{10pt}
  \includegraphics[]{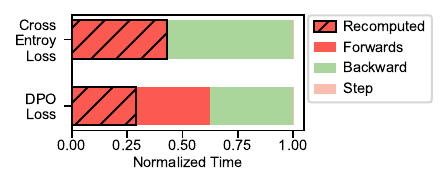}
  \mycaption{fig-train-breakdown}{Continual training time breakdown}{
  When continual training on served inputs, each iteration of training with DPO loss spends 30\% of the total time, shown in hatched red color, recomputing the same activations that have been calculated during serving.
  For pre-training with cross entropy loss, this number increases to 43\%.
  This substantial amount of recomputation time greatly motivates us to reuse the activations that has been calculated during serving. \S \ref{motiv-online-cont-learn-sys}
  }
  \vspace{-10pt}
\end{figure}

However, this schema can be challenging and suboptimal in the design of separated training and serving clusters.
As the model size (\ie, number of parameters) rapidly increases, so does the size of activations as well as the number of activations. 
Attempt to reuse the activations calculated by the serving cluster on a separate training cluster requires transmitting a large amount of data (\ie, activations) over the network and could incur significant transmission overhead. 
It is likely that such overhead can eventually surpass the benefits of reusing the activations (\ie, transmission latency is higher than the recomputation latency). 
Consequently, this design then has to face a dilemma where it has to choose either to transmit activations with high network cost or to recompute the activations with high computation cost. {\em Both} of which are inefficient.

Obviously, a more resource-efficient strategy here is to {\em co-locate} training and serving on the same infrastructure to avoid either transmission or re-computation costs. 
This co-location design enables a powerful key insight: model activations computed during serving can be re-used to reduce or even remove the forward passes in the training loop.

\subsubsection{Harvesting Idle GPU Resource}

In addition to activations reuse, training and serving co-location can harvest the idle resource on the serving cluster.
When allocating resources for application services, it is common practice to {\em overprovision} to ensure that Service Level Objectives (SLOs) are met during unexpected demand spikes. 
In the case of serving LLMs, this conservative measure often results in idle GPU computation cycles and/or memory left on the serving cluster.
By co-locating training with serving, the training job can harvest these idle resources, further helping reduce the cost of online continual learning.

\section{Design Requirements \& Assumption}%
\label{design-req}

Reusing activations for online continual learning and leveraging idle computation cycles is an appealing strategy for cost efficiency. 
However, designing such a system is non-trivial.
Due the intensive computation and large memory requirements from training, the key challenge lies in how to {\em preserve serving latency and capacity} while performing training at the same time.
Hence, we envision that the system should, at a minimum, meet the following design requirements:
\begin{packeditemize}
  \item \textbf{R1}: The system must be able to reuse the activations calculated during serving users to improve the training efficiency compared to a separate training cluster setup.
  \item \textbf{R2}: With activation reuse enabled, the system should have \textbf{minimal or no impact on serving latency}.
  In other words, integrating training into the serving system should not significantly increase the serving SLO violation rate.
  \item \textbf{R3}: Similarly, the system should have \textbf{minimal or no impact on the serving capacity}, meaning the context length and the batch size that originally can be supported should be limited.
\end{packeditemize}
In addition to these requirements, we also make the following assumptions:
\begin{packeditemize}
  \item \textbf{A1}: Due to its computation and memory efficiency, as well as its popularity, Parameter Efficient Fine-Tuning, like LoRA, is practiced on the system.
  \item \textbf{A2}: GPUs' computation and memory resources are overprovisioned to ensure service quality in peak or unexpected spikes in workload.
\end{packeditemize}

\begin{figure}[t]
  \centering
  \includegraphics[width=\linewidth]{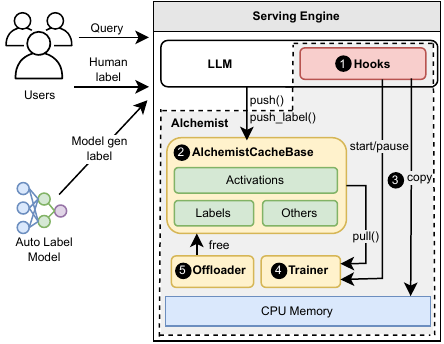}
  \mycaption{fig-system-overview}{\sysname system overview}{
    \tcone \sysname injects preemption hooks to switch from training context to
    serving context upon query arrival.
    \tctwo \sysname saves activations and other cache-able data specified by users calculated during serving jobs for later training when labels are ready.
    \tctri \sysname asynchronously copies serving activations to host memory.
    \tcfour \sysname trainer calls users customized training function which pulls activations and labels when ready.
    \tcfive \sysname frees activations when serving query arrives amid training job and requires more memory. \S\ref{design-intro}
  }
\end{figure}

\section{\sysname Design}%
\label{design-intro}

This section introduces \sysname, the first online continual learning system that efficiently reuses serving activations, to the best of our knowledge. Figure \ref{fig-system-overview} shows the flow and interactions among \sysname's components. In the following sections, we detail how each component meets the design requirements.

\begin{figure}[t]
  \centering
  \hspace{15pt}
  \includegraphics[]{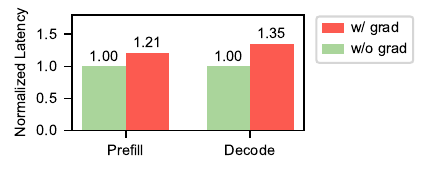}
  \mycaption{fig-grad-overhead}{Latency overhead of activations recording}{
    Due to the cost in recording and saving activations and computation graphs,
    autograd frameworks like \textnormal{\texttt{torch.autograd}} could bring
    up 21\% overhead to the prefill phase and 35\% overhead to each forward
    pass in the decode phase (\ie, 35\% increment in each token's generation
    time). If enabled for each token generated, it will significantly prolong
    the serving latency, violating our latency requirement. \S
    \ref{design-cache}
  }
  \vminten
\end{figure}

\subsection{Record and cache activation}%
\label{design-cache}
To simplify the design process, we set aside \textbf{R3} for now by assuming we have more than enough GPU memory to co-locate serving and training. 
To fulfill \textbf{R1} of reusing serving activations for training, the first step is to record the required activations and cache them for later training use.
Although this may sound straightforward, under the requirements of \textbf{R2}, this can be rather non-trivial. Due to the overhead from the autograd frameworks (\eg, \texttt{torch.enable\_grad}), shown in Figure \ref{fig-grad-overhead}, thoughtlessly enabling activation recording during the generation process can severely impact serving latency, clearly violating \textbf{R2}.

\mypara{Continual pre-train} 
In the case of continual pre-train, model is learning from users' input prompts or the searched context. Hence, loss calculation and backpropagation only require the activations calculated when processing the input (\ie, prefill phase).
This tells that we only need to {\em enable activation recording during prefill phase}.
Since prefill only happens once and is the {\em minority of the total serving time}, such overhead can then be amortized in the end-to-end latency as decode proceeding. 

\mypara{Continual preference alignment}
In the case of continual preference alignment,  this problem can be complicated. 
In preference alignment algorithms like DPO and its other variations, since they directly compare the chosen and rejected responses, it would require the full response from the model to calculate the loss. 
Hence, na\"ively reusing all the activations of the model generated responses (either it is the chosen or the rejected one) would require us to record and save the activations {\em until the end of generation} during serving.
This could introduce prohibitive overhead.

As aforementioned, enabling activation recording could incur up to 35\% latency overhead in the decoding phase.
If we enable it for both prefill and decode phases, this overhead would apply to every forward pass, eventually leading to {\em 35\% latency overhead end-to-end}. 
On the other hand, completely disabling activations recording for both prefill and decode phase to maintain serving latency eliminates the possibility of reusing activations calculated from serving. 
The loss calculation then would require full prefill on the prompt concatenated with chosen and rejected responses {\em twice}, leading to redundant computation. 

Nevertheless, a balanced approach is possible: we can enable activation recording for {\em only the prefill phase}, and \textbf{save the KV cache and the activations} associated with it. 
This design allows loss calculation to only attend to the response texts, skipping the prompts and reusing part of the activations calculated during serving.
Similar to the case of continual pretrain, such overhead then only applies to the prefill phase and will be amortized as generation proceeds.

Since both the chosen response and the rejected response share the same prompt, the two forward passes can share the KV cache as well, further reducing redundant computation. 
Besides the computation saving, sharing KV cache also means sharing the activations associated with it among the two forward passes, hence also reducing unnecessary {\em duplication in memory}.

\subsection{Schedule activation reuse}
\label{design-scheduling}

With the required activations recorded and cached, the next step is to reuse them for training. 
However, simply overlapping training computations with serving computations can severely increase the serving job's latency due to contention, violating the latency requirement specified in \textbf{R2}.

To address this issue, \sysname employs a straightforward yet effective strategy: it \textbf{preempts the training job} and immediately switches to the serving job as soon as a serving query arrives, temporally sharing GPU resources among serving and training jobs.
Leveraging the hook functionality provided by frameworks such as \texttt{PyTorch}, \sysname injects preemption hooks at the start of each layer’s backward function call.
These hooks check for any serving forward pass that is currently running or queued.
If so, the backward pass is paused until the serving forward operations have completed and the queue is empty.

In the case like DPO loss calculation, where both backward passes and forward passes are required during training, similarly, preemption hooks will be injected into each layer's forward.
Through this approach, \sysname minimizes the overlap between training and serving processes, effectively minimizing the impact on serving latency from the training side.
Together with the design in \S\ref{design-cache}, \sysname could fulfill the latency and activation reuse requirement from both \textbf{R1} and \textbf{R2}.

\subsection{\sysname activation offloader}%
\label{design-offload-intro}

While the prior designs address activation reuse requirement from \textbf{R1}
and the latency requirement from \textbf{R2}, another fundamental challenge is
the memory overhead brought by storing the activations when colocating
training with serving. As \textbf{R3} pointed out, \sysname should only bring
minimal or no impact on serving capacity, (\ie, context length and batch
size limits).

However, saving all the activations can risk out-of-memory (OOM). As the context length grows, even if only saving LoRA
adapter's activations and de-duplicate activations by sharing input prompt KV cache as described in \S
\ref{design-cache}, the activations from both the prefill phase and
DPO loss calculation can easily reach 40 GB as shown in Figure \ref{fig-mem-overhead}
when total trained token (\ie, both prompt and responses) is 3000 tokens. If we
save all these activations while serving at the same time, such large memory
overhead can severely limit the context length or the batch size for the
serving side. 
\begin{figure}[t]
  \centering
  \includegraphics[]{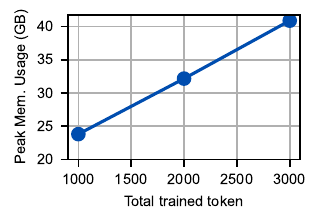}
  \mycaption{fig-mem-overhead}{Peak memory usage with DPO loss}{
    Even if leveraging the input prompt's KV cache as described in
    \S \ref{design-cache}, naively saving all required activations in
    GPU memory on the train side can still introduce significant memory
    overhead. This could severely limit the supported context length or batch
    size on the serving side before OOM. \S \ref{design-offload-intro}.
  }
  \vminten
\end{figure}

Naively offloading every activation to CPU memory, on the other hand, will
leave GPU memory underutilized since serving queries does {\em not
necessarily always} require the full memory space. When offloading more
activations than we actually need, we introduce unnecessary loading
overhead when we load the activations that could have been kept on GPU memory back
to the GPU.

Hence, the goal of \sysname’s offloader is to offload activations \textbf{as
needed} rather than offloading every single activation to CPU memory. To
achieve this, \sysname asynchronously {\em copy} the activations during
the forward pass to a pre-allocated and pinned CPU memory. Unless, even if without overlapping with serving jobs, the activations themselves are too large to be fitted, we will not immediately free the copied activations from GPU memory.
Only when we determine a serving query requires more GPU memory, we then dynamically {\em free} activations in GPU memory until there is enough memory to serve the incoming queries.

Importantly, as physically deallocating GPU memory and reallocating can
be expensive at runtime, the underlying storage buffer for freed
activations is not physically deallocated from the GPU memory, similar to the
memory allocators like \texttt{PyTorch}’s cache allocator~\cite{pytorch_cuda_memory_management}. Instead,
the underlying buffer is marked as \emph{reserved}. When a new tensor needs to
be allocated on the GPU next time, the cache allocator {\em reuses} the
reserved buffer for the new tensor, given the reserved buffer can accommodate
the new tensor. In the case of \sysname, after we free the required number of
activations, the underlying buffer can be immediately reused by the serving
query with no or minimal overhead.

\subsubsection{Offloading map}%
\label{design-offload-map}
To free activations only as needed, we first need to understand {\em how many}
activations we should free. When a new serving query arrives, we see that the
memory consumption --- and whether \sysname risks running out of memory --- is
determined by the following parameters, (1) token
length of cached activations, (2) incoming serving token length, and (3)
incoming serving batch size. As the model size and architecture are known
beforehand, all three parameters have a {\em deterministic mapping} to
the memory consumption. This means, for instance, the memory requirement for
all serving queries with token length $x$ and batch size $y$ remains at a size
of $z$ GB and will not change when the input text or token changes. Similarly,
activations' size for prefilling input with $a$ token length remains at $b$ GB, invariant to the inputs or the tokens.

Hence, before runtime, with a given model and GPU, we can profile a mapping,
namely {\em offloading map}, with these three parameters as input and the
number of bytes to free as the output. In the real system, for the system's
simplicity, we restricted the minimum unit for freeing to {\em one layer's
activations}, instead of bytes or activations. By referring to this profile
mapping, when we start each serving forward at runtime, we may know how many
layers’ activations we must free to avoid the risk of OOM while serving.
In addition, we may also know if there will be OOM {\em during activations recording} due to the size of activations itself. In this case, we directly offload the activations to CPU memory without retaining them in GPU memory at all.

To reduce the profiling effort, during profiling, these three parameters are
incremented by a configurable step size each time. By default, token length of
cached activations and incoming serving are incremented by 500-token steps
while serving batch size is incremented by 5. At runtime, when querying the
mapping, the input value is {\em rounded up} to the nearest recorded step.
For instance, an incoming query with 420 tokens will be rounded to 500 tokens,
as well as its corresponding memory consumption.
Even though this means we always free slightly more than we actually need, it could significantly reduce the offline profiling cost.

\subsubsection{Pipelining}%
\label{design-offload-pipeline}
With offloading map, we know how much we need to offload each time.
The next question is {\em which} activations we should free and {\em which} we should retain in GPU memory.

Since backpropagation proceeds from the output layer to the input layer, the optimal strategy to minimize the wait time is to prioritize {\em retaining the activations of the higher layers} (\ie, layers close to the output layer).
In other words, \sysname should free the activations in the order of forward pass (\ie, from input layer to the output layer).
This ensures that, when backpropagation begins, the gradients for the higher layers can be computed immediately, as their activations remain in GPU memory.
Meanwhile, as backpropagation operates on the layers with activations ready in GPU memory, we can pipeline and prefetch lower layers’ (\ie, layers closest to the input layer) activations.

Ideally, as illustrated in Fig \ref{fig-pipeline}, when backpropagation reaches the layers whose activations were previously freed, the activations have already been loaded back, allowing backpropagation to operate on this layer immediately without waiting for its activations. 
This {\em effectively hides} the delay of loading the activations of these layers.

\begin{figure}[t]
  \centering
  \includegraphics[]{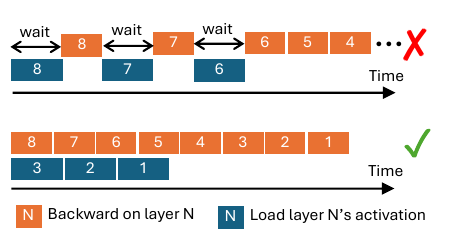}
  \mycaption{fig-pipeline}{\sysname backward and activations loading pipeline}{
    Freeing activations in the forward order and loading them in reverse maximizes the interval between the start of the training job and each layer's backward pass. Since early layers (those closest to the input) are processed last during backpropagation, this approach increases the likelihood that freed activations are reloaded and ready when needed.
   }
\end{figure}

\subsubsection{Hedging map}%
\label{design-offload-hedge}

However, due to the size and the number of activations, it is likely that
computation cannot entirely hide the loading time, even if overlapping
computation and loading. This means that when backpropagation proceeds to the
layers whose activations were freed, those activations may not yet be fully
loaded to GPU, leading to a certain wait time for backpropagation. As the size
and number of activations needed to be loaded increase, the wait time will
prolong as well, and eventually could exceed the time required to recompute
the forward passes from the text input.

To address this issue, we first profile the time required to load activations
back when we free a certain number of layers’ activations. Since the
recomputation time solely depends on the token length and is invariant to the
input content, we can also profile the recomputation time offline. We, then, compare
whether loading time is longer than recomputation time. This allows us to
build another map, named {\em hedging map}, that provides a binary decision
for whether we should load the activations or recompute the forward pass given
the token length of cached activations and the number of layers we freed.

Note here, for loss functions that require multiple forward passes with the same prompt, even if \sysname needs to recompute the forward passes, it still leverages the KV cache sharing technique introduced in \S\ref{design-cache}. This means that it will prefill the user prompts only once and share the corresponding KV cache across multiple forward passes to accelerate the recomputation process and reduce the memory footprint.

\subsubsection{\sysname offloader overall}

To integrate all components in \sysname offloader, right before a new serving
forward starts, \sysname will first determine the new serving forwards’ token
length and batch size, as well as the token length of the cached activations.
Then, it will query the offloading map to determine how many layers’
activations need to be freed. \sysname then will consult the hedging map to
determine, with the given number of layers’ activations to free, if it is more
efficient to load the activations back or just recompute forward passes. If
recomputation is better, \sysname will free all the activations from both GPU
and CPU memory. Otherwise, in the order from the input layer to the output
layer, \sysname will free the given number of layers' activations from GPU
memory. During training \sysname will overlap loading the computation to mask
as much loading time as possible.



\subsection{Cache policy}%
\label{design-cache-eviction}

For simplicity of the system, we currently only allow \sysname to cache one
query's activations. Upon new queries’ arrival, we disable recording and
caching for the new queries rather than evicting the older queries’
activations. This design poses no issue in the case of continual pretrain.
Because continual pretrain only requires the user's prompts or the search
context to start the training job and the activations cached can be
immediately used after the corresponding serving finished.

However, this can be problematic in the case of continual preference
alignment. Continual preference alignment, on the other hand, requires labels
from users’ feedback whose {\em arrival time can be nondeterministic}. It is
possible that a user never submits feedback. The corresponding cached
activation becomes entirely unusable, and the occupied memory space prevents
caching of activations for other queries that have received labels.

To avoid such starvation in caching, we set a configurable timeout threshold
for the cached activations. If cached activations’ label does not arrive
before the timeout, we allow the next serving job’s activations to be
recorded, overwriting the old cached activations.

\section{Implementation}%
\label{implementation}

Following the design depicted in \S\ref{design-intro}, we implemented a prototype of \sysname with 1.5K LOC in Python. 
We rely on commonly used third-party libraries, like \texttt{PyTorch} and Hugging Face's \texttt{transformers}~\cite{transformers.emnlp20} in \sysname's implementation. To ensure \sysname's accessibility and usability, instead of providing it as a standalone serving engine, we implement \sysname as a plugin that can be integrated into existing state-of-the-art serving engines like vLLM~\cite{vllm}, SGLang~\cite{sglang}, \etal.

\mypara{\sysname cache class} 
Different training methods as well as different loss functions may require different inputs that we can cache from serving.
For example, with continual pretrain using cross-entropy loss, the loss function only requires the prefill output and prompt text, but for continual preference alignment with DPO loss, it would require prefill phase KV cache, the accepted response, and rejected response text.
To make \sysname extensible to all possible training methods and loss functions, we provide \texttt{AlchemistCacheBase} abstract class, allowing users to wrap different kinds of cache they would like to store from serving.
In the class, we also implement synchronization functions indicating the label, if needed, for the corresponding cache is ready, for instance, \texttt{is\_ready()} or \texttt{wait\_ready()}.

\mypara{\sysname class}
Users of \sysname are expected to make their model class inherit from \sysname class. This will ensure that, at model initialization, \sysname will register all the scheduling hook functions described in \S\ref{design-scheduling}.
\begin{packeditemize}
    \item \texttt{push(cache: AlchemistCacheBase)}: Within the serving loop, users call this function to push the serving activations and other values wrapped in \texttt{AlchemistCacheBase}.
    \item \texttt{push\_label(label: str)}: If labels are required, users call this function to push the label to \sysname.
    \item \texttt{pull() -> AlchemistCacheBase}: When implementing customized training methods and loss functions, users use this function to pull the cached activations and other values that are required by the training method and loss calculation (\eg, prompt text, label, \etal). 
    This function is blocking as it calls \texttt{AlchemistCacheBase}'s \texttt{wait\_ready()} to wait until the label is ready when it is required.
    \item \texttt{train\_on\_cache()}: This is an abstract function where users can implement their own training logic, which will be called by \sysname on a separate trainer thread.
\end{packeditemize}
With this implementation, besides users' customized training implementation, the users of \sysname are only expected to make {\em minimal changes} to their existing serving engines and model implementations.

\mypara{\sysname offloader}
As a model inheriting from \sysname, we also register \texttt{PyTorch}'s \texttt{saved\_tensor\_hook} \cite{pytorch_autograd_saved_tensors_hooks} to each layer of the models.
This hook function is called each time an activation is recorded by \texttt{PyTorch}'s autograd framework.
We use this hook to (1) asynchronously copy the activations from GPU memory to CPU memory as described in \S\ref{design-offload-intro} and (2) record the ownership of the activations (\ie, which layer's backward requires this activation). 
This ensures we free the correct activations when certain layers' activations are required to be freed. 
We use a CUDA Stream different from \texttt{PyTorch}'s default stream to overlap the computation and data movement between GPU and CPU~\cite{pytorch_cuda_semantics}.
We primarily rely on PyTorch's cache allocator~\cite{pytorch_cuda_memory_management} to virtually free the memory instead of physically de-allocating activations' underlying buffer from GPU memory.
This allows the serving job to immediately reuse the underlying buffer by overwriting the value in the buffer as needed. 

\section{Evaluation}%
\label{eval-intro}

\mypara{Baseline} We evaluate \sysname's performance by comparing it with a separate training cluster that does not reuse serving activations nor share the prompt's KV cache, which serves as our baseline.

\mypara{Hardware} The evaluation testbed is equipped with an Nvidia A100 80GB SXM GPU~\cite{a100gpu} and two AMD EPYC 7V12 64-Core CPUs~\cite{amdcpu}, along with a total of 1.73 TB of CPU memory. 
As offloading may require a relatively large amount of memory, we use \texttt{taskset}~\cite{taskset} to set the evaluation process affinity to one NUMA node to avoid possible NUMA effects.

\mypara{Models} We use Llama-3.1-8B~\cite{llamamodel.corr23}, Mistral-v0.2-Instruct \cite{mistral7b.corr23}, and Phi-4 \cite{phi4.corr25} as our evaluation models. 

\mypara{Dataset} We randomly sampled the prompts from ShareGPT \cite{sharegpt}, which were collected from real users' conversations with ChatGPT, as our input dataset.
We set the output length to be 128 tokens for all queries.
In the case where we evaluate \sysname under various serving loads (\ie, varied queries-per-second, QPS), since ShareGPT does not provide timestamps, following the approach used in many prior works~\cite{vllm, sglang, flexllm}, we generate traces by sampling from a Poisson distribution at varying request rates.

\mypara{Serving} We implement a na\"ive serving engine for fast prototyping and integration with \sysname using Hugging Face.

\mypara{Training} For continual pretrain, we use CrossEntropy loss as an example. For continual preference alignment, we use DPO loss as an example training method.
For DPO loss, we assume model-generated responses are the rejected responses and generate a random tensor with the same length as the chosen response.
We generate the chosen one with minimal delay after the model finishes generating to mimic users providing feedback immediately once they see the model output or auto-labeling LLM.
Models are attached with a LoRA adapter of rank 8 as the example training settings for the evaluation process.

\mypara{Memory profiler settings} For Llama-3.1-8B and Mistral-v0.2-Instruct, we use a step size of 500 tokens for cached token length and incoming query token length and a batch size of 5 when profiling memory consumption. For Phi-4, due to its larger model size, to reduce the profiling error, we decrement the step size to 250 tokens for cached and incoming query.

\mypara{Metrics} We use token-per-second trained to evaluate the training throughput. Since the time \sysname can start its training jobs depending on the actual serving workload, to isolate this factor and focus on \sysname's sole efficiency, we only count the serving idle time when calculating training throughput. When evaluating the serving side, we use time-per-token (TPT) output as the metric.

\begin{figure*}[t]
  \centering
  \includegraphics[]{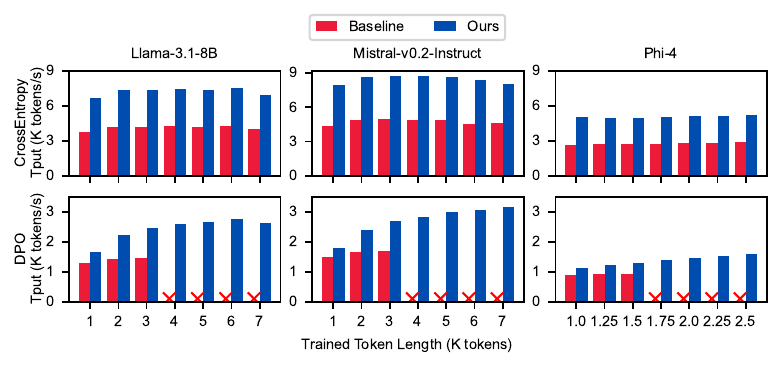}
  \mycaption{fig-tput-improv}{Training throughput improvement}{
    In continual pretrain (CPT) with cross entropy loss (upper figure), \sysname (blue bars) consistently outperforms baseline (red bars) with 1.7x training throughput improvement.
    While in continual preference alignment (CPA) with DPO loss (lower figure), \sysname shows upto 1.68x training throughput improvement before baseline runs out-of-memory (shown with red ``x''). \S\ref{eval-tput-improv}.
  }
\end{figure*}

\subsection{Training throughput improvement}%
\label{eval-tput-improv}
We first evaluate \sysname’s training throughput when {\em no offload is needed} (\ie, serving workload is light).
We sample various lengths of input tokens from the ShareGPT dataset to understand \sysname's performance under various token lengths.
As shown in Figure \ref{fig-tput-improv}), when performing continual pretraining (CPT) with cross-entropy loss (upper figure), \sysname (in blue) consistently outperforms the baseline (in red), delivering over a 1.72x improvement in throughput regardless of token length. For continual preference alignment (CPT) with DPO loss, \sysname achieves between 1.26x and 1.68x improvement, depending on the token length.

This variation arises because the latency of the prefill phase grows linearly—or even super-linearly—with token length. By leveraging the serving layer’s KV cache to bypass the prefill phase during training, \sysname saves more time as the prompt length increases.

Moreover, while the baseline runs out of memory (OOM) for token lengths exceeding 3000 tokens for Llama-3.1-8B and Mistral and 1500 tokens for Phi-4(marked in red ``x''), \sysname can support up to 7000 tokens and 2500 tokens respectively, thanks to the memory savings from reusing the KV cache.

\subsection{Offloading's impact}%
\label{eval-offload}

\begin{figure}[t]
  \centering
  \includegraphics[]{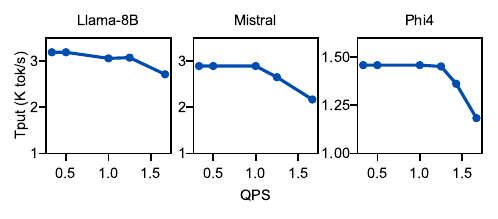}
  \vminfifteen
  \mycaption{fig-offload-qps-vs-tput}{Impact of offloading activations with DPO loss}{
    \sysname's improvement drops as we increase QPS requiring \sysname to offload more activations to CPU memory due to the larger serving batch size and corresponded higher memory requirements.
    This is expected since more activations offloaded means \sysname must load more activations back from CPU memory to GPU memory which can be expensive.
    }
\end{figure}

Next, we evaluate \sysname's performance when offloading is required to accommodate incoming queries' memory requirement due to either larger QPS or longer input token length.
We use \sysname with DPO loss as the evaluation example since DPO loss requires two forward passes hence having a larger memory footprint.
We increase the QPS until there is no idle time for \sysname to run training jobs.
To better illustrate the impact of offloading, when sampling prompts from ShareGPT dataset, we set a minimal token length of 4000 for Llama-3.1-8B and Mistral-v0.2-Instruct and 2500 for Phi4. These numbers were chosen as they are the minimum token length we started to observe the impact of offloading under aforementioned QPS and training settings.

As Figure \ref{fig-offload-qps-vs-tput} shows, as we increase the QPS, due to the larger batch size during serving, \sysname is required to offload more layers to CPU memory. This, as expected, will reduce the training throughput because of the delay when loading the activation back from CPU memory to GPU memory. Especially, in the case of Phi-4, due to its larger model size with higher memory requirements, when at 1.7 QPS, \sysname decides to recompute the majority of the forward passes after querying its hedge mapping as described in \S\ref{design-offload-hedge}.

Since, during recomputation, \sysname still only prefills the input prompt once and shares the KV cache across two responses, it can significantly reduce the memory footprint and avoid OOM, unlike the baseline.
Hence, even with offloading, \sysname is still better than the baseline.

\subsection{\sysname's impact on serving latency}
\label{eval-lat-overhead}
To understand \sysname's impact on the serving side, we start by measuring the TPT distribution under the average request rate of 1.7 queries per second (QPS) with Llama-3.1-8B and DPO loss. 
As the left figure in Fig. \ref{fig-tpt-overhead}, \sysname poses only minimal impact on the TPT distribution and does not introduce large tail latency during serving.

We further evaluate \sysname's impact on the average TPT at various request rates. The result, shown in the right figure in Fig \ref{fig-tpt-overhead}, demonstrates that \sysname increases the average TPT by at most 3\%.

During the attempt to further reduce such overhead, we observe the bulk of the overhead actually comes from the \texttt{PyTorch}'s invocation on the registered hook functions.
In other words, even if we register an empty hook function, such overhead would still incur.
If we are willing to sacrifice the usability of \sysname and add the currently used hook functions as part of models' implementation directly, we can further reduce the latency overhead.

\begin{figure}[t]
  \centering
  \includegraphics[]{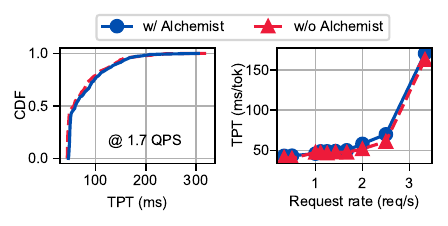}
  \vminten
  \mycaption{fig-tpt-overhead}{\sysname's overhead on serving latency}{The left figure evaluates the distribution of serving time-per-token (TPT) when enable \sysname (blue). 
  The CDF shows that \sysname does not introduce large tail latency and has minimal impact on the TPT's distribution overall compared to disable \sysname (red).
  The right figure compares the average TPT at various request rate.
  The results also indicating \sysname has almost no impact at low request but only introduces at most 3\% overhead at high request when serving system starts to backlog. \S\ref{eval-lat-overhead}.
  }
\end{figure}

\subsection{Memory saving with KV reuse}%
\label{eval-mem-saving}

In the case of continual preference alignment with DPO loss and Llama-3.1-8B model, \sysname leverages the serving prompt's KV cache reuse to reduce the memory duplication of the activations. To evaluate the memory saving, we start by comparing the max token length before OOM. As illustrated in the left figure of Fig. \ref{fig-mem-saving}, without reusing the input prompt's KV cache, baseline can only support up to 3000 tokens with our hardware setup. Meanwhile, with the same setup, by removing the duplication in memory with KV cache reuse, \sysname can support up to 7000 tokens. 

We then directly compare the peak memory usage shown in the right figure of Fig. \ref{fig-mem-saving}. Reusing prompt's KV cache when calculating DPO loss can save 33\% to 47\% peak memory usage depending on the trained token length.

\begin{figure}[t]
  \centering
  \includegraphics[]{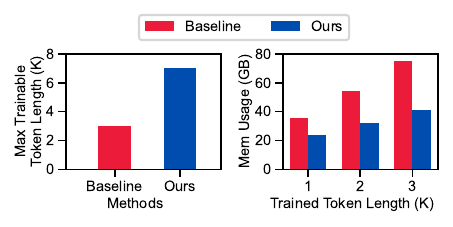}
  \mycaption{fig-mem-saving}{Memory saving when reusing KV cache}{
    By reducing activations duplications, reusing input prompt's KV cache
    during training significantly reduce the peak memory consumption, as much
    as 47\% with 3000 trained token length. With such memory saving, \sysname
    can train 2x longer sample before out-of-memory error. \S\ref{eval-mem-saving}.
  }
\end{figure}

\section{Discussion}%
\label{discussion}

In this paper, \sysname, as an online continual learning system, focuses on addressing the system-side performance issue in reusing the serving activations. However, activation reuse for training may introduce other questions as well. We will address some of the possible questions in this section.

\mypara{Safety of Online Continual Learning at Serving} Obtaining new information and new feedback (either from human users or model) on the updated model implies models continuously trained in online manner are directly deployed to serve future queries.
This could raise concerns about whether the data used for model training is sufficiently safe for deployment.
Fortunately, numerous techniques have been developed to filter training data, ensuring that only safe data points are fed into the training process~\cite{adversarialdpo.corr24, llamaguard.corr23, filtereddpo.corr24, probingtoxiccontent.acl21, beavertails.nips24}.
Additionally, many online learning algorithms already factor in the safety concerns by adding regularization terms or similar paradigms to safeguard continual model learning, ensuring that re-trained models do not drift too far from the pre-trained model while improving generation quality~\cite{onlineaifeedback.corr24, ofsdpo.corr24, learnfromuseredit.nips24, showdonttellaligning.corr24, thingscringeothersiterative.corr24, glam.icml23, sacglam.corr24, reversekl.iclr23}. 

\mypara{Availability in Idle Resource in Serving Workloads} This paper assumes that idle resources in serving workloads allow training jobs to be co-located on the same machine. In practice, serving resources are often over-provisioned to ensure serving quality and capacity under workload spikes. For instance, common autoscaling policies target an average utilization ratio~\cite{k8sautoscale}, typically well below 100\%—often around 60–80\%—to prevent saturation during peak demand.

\mypara{Impact of lower-precision inference} A rising line of work serves LLMs in lower-bit precisions so that the models can fit in GPU memory more easily, and able to hold more inference requests, thus improving the inference throughput~\cite{qserve.corr24, microscopiq.corr24, atom.corr24}. 
A possible concern is whether the activations saved from lower-bits LLMs during serving can be used in training without degrading the quality achieved after training. 
Fortunately, a great many works propose low-bit training that takes in activations with lower precision and shows that it can achieve similar quality as full precision training~\cite{fp4training.corr25, fp8training.corr23, hybridfp8training.nips19}. 

\mypara{Parallelism Strategy Across Training and Serving Processes}
A possible concern is that different optimal parallelism strategies between serving and training may impact the training throughput when co-locating serving and training. We leave this as future work.

\section{Related Works}%
\label{related}

\mypara{Serving systems}
One of the most popular lines of LLM systems research focuses on developing
efficient serving systems for LLMs, to improve the serving
throughput~\cite{vllm, sarathi, orca, preble, distserve, splitwise, parrot,
  cao2025localityawarefairschedulingllm, vllm}, or reduce the inference
latency~\cite{cachegen, cacheblend, specinfer, conserve, flashattention}.
Common optimizations include caching and reusing the KV cache of other
prompts~\cite{cachegen, attentionstore, ragcache}, offloading KV cache to CPU
or storage devices~\cite{infinigen, cachegen, mooncake.fast25}, better parallelization
strategies that minimize network communication overhead~\cite{distserve, mooncake.fast25,
  microsoft_mscclpp}, and faster attention CUDA kernels~\cite{flashattention, flashinfer.corr25, xFormers2022}.
\sysname is orthogonal to this line of work. The above optimizations
developed by previous work can be easily adapted to \sysname to make its
inference part faster and more efficient.

\mypara{Training systems}
Another line of LLM system work aims at designing efficient systems for LLM
training. Some works improve parallelism strategies, such as designing
different distributed parallelism for different submodules in multi-modal
model training~\cite{Huang2024}, designing dynamic parallel strategies
according to input sequence
length~\cite{li2024demystifyingworkloadimbalanceslarge}, or reducing idle
bubbles in pipeline parallelism~\cite{pipelineparallelism.corr19}. Other works improve
fault tolerance in distributed training by efficient model
checkpointing~\cite{lian2024universalcheckpointingefficientflexible, wan2024bytecheckpointunifiedcheckpointinglarge}, or quickly detecting faulty machines in
distributed training~\cite{deng2024minderfaultymachinedetection} and
fastly recovering from failure~\cite{recover}. Finally, an emerging line of
works builds faster training frameworks for RLHF, such as employing intra-stage
and inter-stage fusion to improve GPU
utilization~\cite{zhong2024rlhfuseefficientrlhftraining}, or designing APIs
for decoupling and orchestrating computation and data dependencies in RLHF
dataflows.

\sysname is also complementary to this line of work, as the optimizations they
bring up can be applied to the training part of \sysname to make its training
faster or higher throughput.

\mypara{GPU multiplexing for training and serving}
Efforts on co-locating inference and training in video analytics relate well
to \sysname. Specifically, Ekya and RECL~\cite{ekya, recl} enable continuous
retraining on new video samples, and jointly host it with inference on the
same device by letting them share GPU resources with technologies like Nvidia MPS~\cite{nvidia-mps}. Co-locating inference and
training in LLMs is much harder than that for video analytics because these
works assume the workload is deterministic. For example, the resolution and
frame rate of the incoming stream are known. Hence, Ekya and RECL understand  
how much resource (\eg, memory and computation cycles) the training side
should occupy.

A more related work, flexLLM~\cite{flexllm}, co-locates LLM inference and LoRA
fine-tuning by fusing serving and training kernels, effectively sharing the GPU's computation unit spatially. However, it only considers offline fine-tuning on different input
data than that to the serving jobs, thus failing to reuse activations from
serving, leading to redundant forward recomputation. This work can be complementary to \sysname as its techniques in kernel fusion can further help \sysname harvesting the idle computation units even when serving jobs are running.

\mypara{Continuous learning}
Continual learning for large language models has rapidly evolved, with works like Ernie \cite{ernie.aaai20} and Don't Stop Pretraining \cite{dontstoppretrainingadapt.corr20} demonstrating that continuous pre-training enables models to integrate new domain-specific knowledge without forgetting previous capabilities. Complementary approaches --- ranging from reformulated domain adaptation frameworks \cite{reformulating.corr23} and reading comprehension-based domain tuning \cite{adaptinglargelanguagemodels.corr24} to strategies for mitigating forgetting across modalities \cite{continualpretrainingmitigatesforgetting.corr22} --- have further highlighted the effectiveness of incremental learning.

In parallel, research has increasingly focused on aligning LLM outputs with human preferences and values. Studies on fine-tuning from human feedback \cite{finetuninglanguagemodelshuman.corr20, learningsummarizehumanfeedback.corr22, dpo.corr24, ppo.corr17}, optimal policy fitting \cite{copf.corr23}, and f-divergence minimization for preference alignment \cite{aligninglanguagemodelspreferences.corr23} underscore the importance of ethical and user-aligned model adaptation. Moreover, parameter-efficient tuning techniques such as multi-task prompt tuning \cite{multitaskprompttuning.corr23} and ConPET \cite{conpet.corr23} offer promising solutions to balance continuous learning demands with computational efficiency.

\vminfive
\section{Conclusion}%
\label{conclude}

 By reusing intermediate activations produced during serving, \sysname bridges the gap between real-time application demands and resource-intensive model updates. This approach not only increases training throughput by up to 1.72x and reduces memory overhead by as much as 47\%, but it also maintains the stringent latency requirements essential for high-quality service delivery.

These findings underscore the potential of reusing serving activations to unlock performance gains in online continual learning scenarios. Moving forward, further optimization of gradient recording techniques and dynamic activation management could pave the way for even more efficient LLM systems, ultimately enabling more scalable AI services.

\bibliographystyle{ACM-Reference-Format}

\bibliography{local.bib}


\begin{thebibliography}{95}


\ifx \showCODEN    \undefined \def \showCODEN     #1{\unskip}     \fi
\ifx \showISBNx    \undefined \def \showISBNx     #1{\unskip}     \fi
\ifx \showISBNxiii \undefined \def \showISBNxiii  #1{\unskip}     \fi
\ifx \showISSN     \undefined \def \showISSN      #1{\unskip}     \fi
\ifx \showLCCN     \undefined \def \showLCCN      #1{\unskip}     \fi
\ifx \shownote     \undefined \def \shownote      #1{#1}          \fi
\ifx \showarticletitle \undefined \def \showarticletitle #1{#1}   \fi
\ifx \showURL      \undefined \def \showURL       {\relax}        \fi
\providecommand\bibfield[2]{#2}
\providecommand\bibinfo[2]{#2}
\providecommand\natexlab[1]{#1}
\providecommand\showeprint[2][]{arXiv:#2}

\bibitem[nvi(2025)]%
        {nvidia-mps}
 \bibinfo{year}{2025}\natexlab{}.
\newblock \bibinfo{title}{Multi-Process Service}.
\newblock \bibinfo{howpublished}{\url{https://docs.nvidia.com/deploy/mps/index.html}}.
\newblock
\newblock
\shownote{NVIDIA Documentation}.


\bibitem[Abdin et~al\mbox{.}(2024)]%
        {phi4.corr25}
\bibfield{author}{\bibinfo{person}{Marah Abdin}, \bibinfo{person}{Jyoti Aneja}, \bibinfo{person}{Harkirat Behl}, \bibinfo{person}{Sébastien Bubeck}, \bibinfo{person}{Ronen Eldan}, \bibinfo{person}{Suriya Gunasekar}, \bibinfo{person}{Michael Harrison}, \bibinfo{person}{Russell~J. Hewett}, \bibinfo{person}{Mojan Javaheripi}, \bibinfo{person}{Piero Kauffmann}, \bibinfo{person}{James~R. Lee}, \bibinfo{person}{Yin~Tat Lee}, \bibinfo{person}{Yuanzhi Li}, \bibinfo{person}{Weishung Liu}, \bibinfo{person}{Caio C.~T. Mendes}, \bibinfo{person}{Anh Nguyen}, \bibinfo{person}{Eric Price}, \bibinfo{person}{Gustavo de Rosa}, \bibinfo{person}{Olli Saarikivi}, \bibinfo{person}{Adil Salim}, \bibinfo{person}{Shital Shah}, \bibinfo{person}{Xin Wang}, \bibinfo{person}{Rachel Ward}, \bibinfo{person}{Yue Wu}, \bibinfo{person}{Dingli Yu}, \bibinfo{person}{Cyril Zhang}, {and} \bibinfo{person}{Yi Zhang}.} \bibinfo{year}{2024}\natexlab{}.
\newblock \bibinfo{title}{Phi-4 Technical Report}.
\newblock
\showeprint[arxiv]{2412.08905}~[cs.CL]
\urldef\tempurl%
\url{https://arxiv.org/abs/2412.08905}
\showURL{%
\tempurl}


\bibitem[{AMD}(2025)]%
        {amdcpu}
\bibfield{author}{\bibinfo{person}{{AMD}}.} \bibinfo{year}{2025}\natexlab{}.
\newblock \bibinfo{title}{AMD EPYC™ 7002 Series Processors}.
\newblock \bibinfo{howpublished}{\url{https://www.amd.com/en/products/processors/server/epyc/7002-series.html}}.
\newblock
\newblock
\shownote{Accessed: 2025-02-17}.


\bibitem[{anon8231489123}(2025)]%
        {sharegpt}
\bibfield{author}{\bibinfo{person}{{anon8231489123}}.} \bibinfo{year}{2025}\natexlab{}.
\newblock \bibinfo{title}{ShareGPT Vicuna Unfiltered}.
\newblock \bibinfo{howpublished}{\url{https://huggingface.co/datasets/anon8231489123/ShareGPT_Vicuna_unfiltered}}.
\newblock
\newblock
\shownote{Hugging Face Dataset. Accessed: 2025-02-14}.


\bibitem[Anonymous(2024)]%
        {sarathi}
\bibfield{author}{\bibinfo{person}{Anonymous}.} \bibinfo{year}{2024}\natexlab{}.
\newblock \bibinfo{title}{ChunkAttention: Efficient Attention on {KV} Cache with Chunking Sharing and Batching}.
\newblock
\urldef\tempurl%
\url{https://openreview.net/forum?id=9k27IITeAZ}
\showURL{%
\tempurl}


\bibitem[Askell et~al\mbox{.}(2021)]%
        {generallanguageassistantlaboratory.corr21}
\bibfield{author}{\bibinfo{person}{Amanda Askell}, \bibinfo{person}{Yuntao Bai}, \bibinfo{person}{Anna Chen}, \bibinfo{person}{Dawn Drain}, \bibinfo{person}{Deep Ganguli}, \bibinfo{person}{Tom Henighan}, \bibinfo{person}{Andy Jones}, \bibinfo{person}{Nicholas Joseph}, \bibinfo{person}{Ben Mann}, \bibinfo{person}{Nova DasSarma}, \bibinfo{person}{Nelson Elhage}, \bibinfo{person}{Zac Hatfield-Dodds}, \bibinfo{person}{Danny Hernandez}, \bibinfo{person}{Jackson Kernion}, \bibinfo{person}{Kamal Ndousse}, \bibinfo{person}{Catherine Olsson}, \bibinfo{person}{Dario Amodei}, \bibinfo{person}{Tom Brown}, \bibinfo{person}{Jack Clark}, \bibinfo{person}{Sam McCandlish}, \bibinfo{person}{Chris Olah}, {and} \bibinfo{person}{Jared Kaplan}.} \bibinfo{year}{2021}\natexlab{}.
\newblock \bibinfo{title}{A General Language Assistant as a Laboratory for Alignment}.
\newblock
\showeprint[arxiv]{2112.00861}~[cs.CL]
\urldef\tempurl%
\url{https://arxiv.org/abs/2112.00861}
\showURL{%
\tempurl}


\bibitem[Bhardwaj et~al\mbox{.}(2020)]%
        {ekya}
\bibfield{author}{\bibinfo{person}{Romil Bhardwaj}, \bibinfo{person}{Zhengxu Xia}, \bibinfo{person}{Ganesh Ananthanarayanan}, \bibinfo{person}{Junchen Jiang}, \bibinfo{person}{Nikolaos Karianakis}, \bibinfo{person}{Yuanchao Shu}, \bibinfo{person}{Kevin Hsieh}, \bibinfo{person}{Victor Bahl}, {and} \bibinfo{person}{Ion Stoica}.} \bibinfo{year}{2020}\natexlab{}.
\newblock \bibinfo{title}{Ekya: Continuous Learning of Video Analytics Models on Edge Compute Servers}.
\newblock
\showeprint[arxiv]{2012.10557}~[cs.DC]
\urldef\tempurl%
\url{https://arxiv.org/abs/2012.10557}
\showURL{%
\tempurl}


\bibitem[Blau et~al\mbox{.}(2024)]%
        {cpt.corr24}
\bibfield{author}{\bibinfo{person}{Tsachi Blau}, \bibinfo{person}{Moshe Kimhi}, \bibinfo{person}{Yonatan Belinkov}, \bibinfo{person}{Alexander Bronstein}, {and} \bibinfo{person}{Chaim Baskin}.} \bibinfo{year}{2024}\natexlab{}.
\newblock \bibinfo{title}{Context-aware Prompt Tuning: Advancing In-Context Learning with Adversarial Methods}.
\newblock
\showeprint[arxiv]{2410.17222}~[cs.CL]
\urldef\tempurl%
\url{https://arxiv.org/abs/2410.17222}
\showURL{%
\tempurl}


\bibitem[Cao et~al\mbox{.}(2025)]%
        {cao2025localityawarefairschedulingllm}
\bibfield{author}{\bibinfo{person}{Shiyi Cao}, \bibinfo{person}{Yichuan Wang}, \bibinfo{person}{Ziming Mao}, \bibinfo{person}{Pin-Lun Hsu}, \bibinfo{person}{Liangsheng Yin}, \bibinfo{person}{Tian Xia}, \bibinfo{person}{Dacheng Li}, \bibinfo{person}{Shu Liu}, \bibinfo{person}{Yineng Zhang}, \bibinfo{person}{Yang Zhou}, \bibinfo{person}{Ying Sheng}, \bibinfo{person}{Joseph Gonzalez}, {and} \bibinfo{person}{Ion Stoica}.} \bibinfo{year}{2025}\natexlab{}.
\newblock \bibinfo{title}{Locality-aware Fair Scheduling in LLM Serving}.
\newblock
\showeprint[arxiv]{2501.14312}~[cs.DC]
\urldef\tempurl%
\url{https://arxiv.org/abs/2501.14312}
\showURL{%
\tempurl}


\bibitem[Carta et~al\mbox{.}(2024)]%
        {glam.icml23}
\bibfield{author}{\bibinfo{person}{Thomas Carta}, \bibinfo{person}{Clément Romac}, \bibinfo{person}{Thomas Wolf}, \bibinfo{person}{Sylvain Lamprier}, \bibinfo{person}{Olivier Sigaud}, {and} \bibinfo{person}{Pierre-Yves Oudeyer}.} \bibinfo{year}{2024}\natexlab{}.
\newblock \bibinfo{title}{Grounding Large Language Models in Interactive Environments with Online Reinforcement Learning}.
\newblock
\showeprint[arxiv]{2302.02662}~[cs.LG]
\urldef\tempurl%
\url{https://arxiv.org/abs/2302.02662}
\showURL{%
\tempurl}


\bibitem[Cheng et~al\mbox{.}(2024)]%
        {adaptinglargelanguagemodels.corr24}
\bibfield{author}{\bibinfo{person}{Daixuan Cheng}, \bibinfo{person}{Shaohan Huang}, {and} \bibinfo{person}{Furu Wei}.} \bibinfo{year}{2024}\natexlab{}.
\newblock \bibinfo{title}{Adapting Large Language Models to Domains via Reading Comprehension}.
\newblock
\showeprint[arxiv]{2309.09530}~[cs.CL]
\urldef\tempurl%
\url{https://arxiv.org/abs/2309.09530}
\showURL{%
\tempurl}


\bibitem[Company(2023)]%
        {mckinsey2023ai}
\bibfield{author}{\bibinfo{person}{McKinsey~\& Company}.} \bibinfo{year}{2023}\natexlab{}.
\newblock \bibinfo{title}{The State of AI in 2023: Generative AI’s Breakout Year}.
\newblock
\urldef\tempurl%
\url{https://www.mckinsey.com}
\showURL{%
\tempurl}


\bibitem[Cossu et~al\mbox{.}(2022)]%
        {continualpretrainingmitigatesforgetting.corr22}
\bibfield{author}{\bibinfo{person}{Andrea Cossu}, \bibinfo{person}{Tinne Tuytelaars}, \bibinfo{person}{Antonio Carta}, \bibinfo{person}{Lucia Passaro}, \bibinfo{person}{Vincenzo Lomonaco}, {and} \bibinfo{person}{Davide Bacciu}.} \bibinfo{year}{2022}\natexlab{}.
\newblock \bibinfo{title}{Continual Pre-Training Mitigates Forgetting in Language and Vision}.
\newblock
\showeprint[arxiv]{2205.09357}~[cs.LG]
\urldef\tempurl%
\url{https://arxiv.org/abs/2205.09357}
\showURL{%
\tempurl}


\bibitem[Dao et~al\mbox{.}(2022)]%
        {flashattention}
\bibfield{author}{\bibinfo{person}{Tri Dao}, \bibinfo{person}{Daniel~Y. Fu}, \bibinfo{person}{Stefano Ermon}, \bibinfo{person}{Atri Rudra}, {and} \bibinfo{person}{Christopher Ré}.} \bibinfo{year}{2022}\natexlab{}.
\newblock \bibinfo{title}{{FlashAttention: Fast and Memory-Efficient Exact Attention with IO-Awareness}}.
\newblock
\showeprint[arxiv]{2205.14135}~[cs.LG]


\bibitem[DeepSeek-AI et~al\mbox{.}(2025)]%
        {deepseek}
\bibfield{author}{\bibinfo{person}{DeepSeek-AI}, \bibinfo{person}{Daya Guo}, \bibinfo{person}{Dejian Yang}, \bibinfo{person}{Haowei Zhang}, \bibinfo{person}{Junxiao Song}, \bibinfo{person}{Ruoyu Zhang}, \bibinfo{person}{Runxin Xu}, \bibinfo{person}{Qihao Zhu}, \bibinfo{person}{Shirong Ma}, \bibinfo{person}{Peiyi Wang}, \bibinfo{person}{Xiao Bi}, \bibinfo{person}{Xiaokang Zhang}, \bibinfo{person}{Xingkai Yu}, \bibinfo{person}{Yu Wu}, \bibinfo{person}{Z.~F. Wu}, \bibinfo{person}{Zhibin Gou}, \bibinfo{person}{Zhihong Shao}, \bibinfo{person}{Zhuoshu Li}, \bibinfo{person}{Ziyi Gao}, \bibinfo{person}{Aixin Liu}, \bibinfo{person}{Bing Xue}, \bibinfo{person}{Bingxuan Wang}, \bibinfo{person}{Bochao Wu}, \bibinfo{person}{Bei Feng}, \bibinfo{person}{Chengda Lu}, \bibinfo{person}{Chenggang Zhao}, \bibinfo{person}{Chengqi Deng}, \bibinfo{person}{Chenyu Zhang}, \bibinfo{person}{Chong Ruan}, \bibinfo{person}{Damai Dai}, \bibinfo{person}{Deli Chen}, \bibinfo{person}{Dongjie Ji}, \bibinfo{person}{Erhang Li},
  \bibinfo{person}{Fangyun Lin}, \bibinfo{person}{Fucong Dai}, \bibinfo{person}{Fuli Luo}, \bibinfo{person}{Guangbo Hao}, \bibinfo{person}{Guanting Chen}, \bibinfo{person}{Guowei Li}, \bibinfo{person}{H. Zhang}, \bibinfo{person}{Han Bao}, \bibinfo{person}{Hanwei Xu}, \bibinfo{person}{Haocheng Wang}, \bibinfo{person}{Honghui Ding}, \bibinfo{person}{Huajian Xin}, \bibinfo{person}{Huazuo Gao}, \bibinfo{person}{Hui Qu}, \bibinfo{person}{Hui Li}, \bibinfo{person}{Jianzhong Guo}, \bibinfo{person}{Jiashi Li}, \bibinfo{person}{Jiawei Wang}, \bibinfo{person}{Jingchang Chen}, \bibinfo{person}{Jingyang Yuan}, \bibinfo{person}{Junjie Qiu}, \bibinfo{person}{Junlong Li}, \bibinfo{person}{J.~L. Cai}, \bibinfo{person}{Jiaqi Ni}, \bibinfo{person}{Jian Liang}, \bibinfo{person}{Jin Chen}, \bibinfo{person}{Kai Dong}, \bibinfo{person}{Kai Hu}, \bibinfo{person}{Kaige Gao}, \bibinfo{person}{Kang Guan}, \bibinfo{person}{Kexin Huang}, \bibinfo{person}{Kuai Yu}, \bibinfo{person}{Lean Wang}, \bibinfo{person}{Lecong Zhang},
  \bibinfo{person}{Liang Zhao}, \bibinfo{person}{Litong Wang}, \bibinfo{person}{Liyue Zhang}, \bibinfo{person}{Lei Xu}, \bibinfo{person}{Leyi Xia}, \bibinfo{person}{Mingchuan Zhang}, \bibinfo{person}{Minghua Zhang}, \bibinfo{person}{Minghui Tang}, \bibinfo{person}{Meng Li}, \bibinfo{person}{Miaojun Wang}, \bibinfo{person}{Mingming Li}, \bibinfo{person}{Ning Tian}, \bibinfo{person}{Panpan Huang}, \bibinfo{person}{Peng Zhang}, \bibinfo{person}{Qiancheng Wang}, \bibinfo{person}{Qinyu Chen}, \bibinfo{person}{Qiushi Du}, \bibinfo{person}{Ruiqi Ge}, \bibinfo{person}{Ruisong Zhang}, \bibinfo{person}{Ruizhe Pan}, \bibinfo{person}{Runji Wang}, \bibinfo{person}{R.~J. Chen}, \bibinfo{person}{R.~L. Jin}, \bibinfo{person}{Ruyi Chen}, \bibinfo{person}{Shanghao Lu}, \bibinfo{person}{Shangyan Zhou}, \bibinfo{person}{Shanhuang Chen}, \bibinfo{person}{Shengfeng Ye}, \bibinfo{person}{Shiyu Wang}, \bibinfo{person}{Shuiping Yu}, \bibinfo{person}{Shunfeng Zhou}, \bibinfo{person}{Shuting Pan}, \bibinfo{person}{S.~S. Li},
  \bibinfo{person}{Shuang Zhou}, \bibinfo{person}{Shaoqing Wu}, \bibinfo{person}{Shengfeng Ye}, \bibinfo{person}{Tao Yun}, \bibinfo{person}{Tian Pei}, \bibinfo{person}{Tianyu Sun}, \bibinfo{person}{T. Wang}, \bibinfo{person}{Wangding Zeng}, \bibinfo{person}{Wanjia Zhao}, \bibinfo{person}{Wen Liu}, \bibinfo{person}{Wenfeng Liang}, \bibinfo{person}{Wenjun Gao}, \bibinfo{person}{Wenqin Yu}, \bibinfo{person}{Wentao Zhang}, \bibinfo{person}{W.~L. Xiao}, \bibinfo{person}{Wei An}, \bibinfo{person}{Xiaodong Liu}, \bibinfo{person}{Xiaohan Wang}, \bibinfo{person}{Xiaokang Chen}, \bibinfo{person}{Xiaotao Nie}, \bibinfo{person}{Xin Cheng}, \bibinfo{person}{Xin Liu}, \bibinfo{person}{Xin Xie}, \bibinfo{person}{Xingchao Liu}, \bibinfo{person}{Xinyu Yang}, \bibinfo{person}{Xinyuan Li}, \bibinfo{person}{Xuecheng Su}, \bibinfo{person}{Xuheng Lin}, \bibinfo{person}{X.~Q. Li}, \bibinfo{person}{Xiangyue Jin}, \bibinfo{person}{Xiaojin Shen}, \bibinfo{person}{Xiaosha Chen}, \bibinfo{person}{Xiaowen Sun}, \bibinfo{person}{Xiaoxiang
  Wang}, \bibinfo{person}{Xinnan Song}, \bibinfo{person}{Xinyi Zhou}, \bibinfo{person}{Xianzu Wang}, \bibinfo{person}{Xinxia Shan}, \bibinfo{person}{Y.~K. Li}, \bibinfo{person}{Y.~Q. Wang}, \bibinfo{person}{Y.~X. Wei}, \bibinfo{person}{Yang Zhang}, \bibinfo{person}{Yanhong Xu}, \bibinfo{person}{Yao Li}, \bibinfo{person}{Yao Zhao}, \bibinfo{person}{Yaofeng Sun}, \bibinfo{person}{Yaohui Wang}, \bibinfo{person}{Yi Yu}, \bibinfo{person}{Yichao Zhang}, \bibinfo{person}{Yifan Shi}, \bibinfo{person}{Yiliang Xiong}, \bibinfo{person}{Ying He}, \bibinfo{person}{Yishi Piao}, \bibinfo{person}{Yisong Wang}, \bibinfo{person}{Yixuan Tan}, \bibinfo{person}{Yiyang Ma}, \bibinfo{person}{Yiyuan Liu}, \bibinfo{person}{Yongqiang Guo}, \bibinfo{person}{Yuan Ou}, \bibinfo{person}{Yuduan Wang}, \bibinfo{person}{Yue Gong}, \bibinfo{person}{Yuheng Zou}, \bibinfo{person}{Yujia He}, \bibinfo{person}{Yunfan Xiong}, \bibinfo{person}{Yuxiang Luo}, \bibinfo{person}{Yuxiang You}, \bibinfo{person}{Yuxuan Liu}, \bibinfo{person}{Yuyang Zhou},
  \bibinfo{person}{Y.~X. Zhu}, \bibinfo{person}{Yanhong Xu}, \bibinfo{person}{Yanping Huang}, \bibinfo{person}{Yaohui Li}, \bibinfo{person}{Yi Zheng}, \bibinfo{person}{Yuchen Zhu}, \bibinfo{person}{Yunxian Ma}, \bibinfo{person}{Ying Tang}, \bibinfo{person}{Yukun Zha}, \bibinfo{person}{Yuting Yan}, \bibinfo{person}{Z.~Z. Ren}, \bibinfo{person}{Zehui Ren}, \bibinfo{person}{Zhangli Sha}, \bibinfo{person}{Zhe Fu}, \bibinfo{person}{Zhean Xu}, \bibinfo{person}{Zhenda Xie}, \bibinfo{person}{Zhengyan Zhang}, \bibinfo{person}{Zhewen Hao}, \bibinfo{person}{Zhicheng Ma}, \bibinfo{person}{Zhigang Yan}, \bibinfo{person}{Zhiyu Wu}, \bibinfo{person}{Zihui Gu}, \bibinfo{person}{Zijia Zhu}, \bibinfo{person}{Zijun Liu}, \bibinfo{person}{Zilin Li}, \bibinfo{person}{Ziwei Xie}, \bibinfo{person}{Ziyang Song}, \bibinfo{person}{Zizheng Pan}, \bibinfo{person}{Zhen Huang}, \bibinfo{person}{Zhipeng Xu}, \bibinfo{person}{Zhongyu Zhang}, {and} \bibinfo{person}{Zhen Zhang}.} \bibinfo{year}{2025}\natexlab{}.
\newblock \bibinfo{title}{DeepSeek-R1: Incentivizing Reasoning Capability in LLMs via Reinforcement Learning}.
\newblock
\showeprint[arxiv]{2501.12948}~[cs.CL]
\urldef\tempurl%
\url{https://arxiv.org/abs/2501.12948}
\showURL{%
\tempurl}


\bibitem[Deng et~al\mbox{.}(2024)]%
        {deng2024minderfaultymachinedetection}
\bibfield{author}{\bibinfo{person}{Yangtao Deng}, \bibinfo{person}{Xiang Shi}, \bibinfo{person}{Zhuo Jiang}, \bibinfo{person}{Xingjian Zhang}, \bibinfo{person}{Lei Zhang}, \bibinfo{person}{Zhang Zhang}, \bibinfo{person}{Bo Li}, \bibinfo{person}{Zuquan Song}, \bibinfo{person}{Hang Zhu}, \bibinfo{person}{Gaohong Liu}, \bibinfo{person}{Fuliang Li}, \bibinfo{person}{Shuguang Wang}, \bibinfo{person}{Haibin Lin}, \bibinfo{person}{Jianxi Ye}, {and} \bibinfo{person}{Minlan Yu}.} \bibinfo{year}{2024}\natexlab{}.
\newblock \bibinfo{title}{Minder: Faulty Machine Detection for Large-scale Distributed Model Training}.
\newblock
\showeprint[arxiv]{2411.01791}~[cs.DC]
\urldef\tempurl%
\url{https://arxiv.org/abs/2411.01791}
\showURL{%
\tempurl}


\bibitem[Gabriel(2020)]%
        {Gabriel_2020}
\bibfield{author}{\bibinfo{person}{Iason Gabriel}.} \bibinfo{year}{2020}\natexlab{}.
\newblock \showarticletitle{Artificial Intelligence, Values, and Alignment}.
\newblock \bibinfo{journal}{\emph{Minds and Machines}} \bibinfo{volume}{30}, \bibinfo{number}{3} (\bibinfo{date}{Sept.} \bibinfo{year}{2020}), \bibinfo{pages}{411–437}.
\newblock
\showISSN{1572-8641}
\href{https://doi.org/10.1007/s11023-020-09539-2}{doi:\nolinkurl{10.1007/s11023-020-09539-2}}


\bibitem[Gao et~al\mbox{.}(2024a)]%
        {attentionstore}
\bibfield{author}{\bibinfo{person}{Bin Gao}, \bibinfo{person}{Zhuomin He}, \bibinfo{person}{Puru Sharma}, \bibinfo{person}{Qingxuan Kang}, \bibinfo{person}{Djordje Jevdjic}, \bibinfo{person}{Junbo Deng}, \bibinfo{person}{Xingkun Yang}, \bibinfo{person}{Zhou Yu}, {and} \bibinfo{person}{Pengfei Zuo}.} \bibinfo{year}{2024}\natexlab{a}.
\newblock \showarticletitle{AttentionStore: Cost-effective Attention Reuse across Multi-turn Conversations in Large Language Model Serving}.
\newblock \bibinfo{journal}{\emph{arXiv preprint arXiv:2403.19708}} (\bibinfo{year}{2024}).
\newblock


\bibitem[Gao et~al\mbox{.}(2024b)]%
        {learnfromuseredit.nips24}
\bibfield{author}{\bibinfo{person}{Ge Gao}, \bibinfo{person}{Alexey Taymanov}, \bibinfo{person}{Eduardo Salinas}, \bibinfo{person}{Paul Mineiro}, {and} \bibinfo{person}{Dipendra Misra}.} \bibinfo{year}{2024}\natexlab{b}.
\newblock \showarticletitle{Aligning llm agents by learning latent preference from user edits}.
\newblock \bibinfo{journal}{\emph{arXiv preprint arXiv:2404.15269}} (\bibinfo{year}{2024}).
\newblock


\bibitem[Gaven et~al\mbox{.}(2024)]%
        {sacglam.corr24}
\bibfield{author}{\bibinfo{person}{Loris Gaven}, \bibinfo{person}{Clement Romac}, \bibinfo{person}{Thomas Carta}, \bibinfo{person}{Sylvain Lamprier}, \bibinfo{person}{Olivier Sigaud}, {and} \bibinfo{person}{Pierre-Yves Oudeyer}.} \bibinfo{year}{2024}\natexlab{}.
\newblock \bibinfo{title}{SAC-GLAM: Improving Online RL for LLM agents with Soft Actor-Critic and Hindsight Relabeling}.
\newblock
\showeprint[arxiv]{2410.12481}~[cs.LG]
\urldef\tempurl%
\url{https://arxiv.org/abs/2410.12481}
\showURL{%
\tempurl}


\bibitem[Go et~al\mbox{.}(2023)]%
        {aligninglanguagemodelspreferences.corr23}
\bibfield{author}{\bibinfo{person}{Dongyoung Go}, \bibinfo{person}{Tomasz Korbak}, \bibinfo{person}{Germán Kruszewski}, \bibinfo{person}{Jos Rozen}, \bibinfo{person}{Nahyeon Ryu}, {and} \bibinfo{person}{Marc Dymetman}.} \bibinfo{year}{2023}\natexlab{}.
\newblock \bibinfo{title}{Aligning Language Models with Preferences through f-divergence Minimization}.
\newblock
\showeprint[arxiv]{2302.08215}~[cs.CL]
\urldef\tempurl%
\url{https://arxiv.org/abs/2302.08215}
\showURL{%
\tempurl}


\bibitem[Guo et~al\mbox{.}(2024)]%
        {onlineaifeedback.corr24}
\bibfield{author}{\bibinfo{person}{Shangmin Guo}, \bibinfo{person}{Biao Zhang}, \bibinfo{person}{Tianlin Liu}, \bibinfo{person}{Tianqi Liu}, \bibinfo{person}{Misha Khalman}, \bibinfo{person}{Felipe Llinares}, \bibinfo{person}{Alexandre Rame}, \bibinfo{person}{Thomas Mesnard}, \bibinfo{person}{Yao Zhao}, \bibinfo{person}{Bilal Piot}, \bibinfo{person}{Johan Ferret}, {and} \bibinfo{person}{Mathieu Blondel}.} \bibinfo{year}{2024}\natexlab{}.
\newblock \bibinfo{title}{Direct Language Model Alignment from Online AI Feedback}.
\newblock
\showeprint[arxiv]{2402.04792}~[cs.AI]
\urldef\tempurl%
\url{https://arxiv.org/abs/2402.04792}
\showURL{%
\tempurl}


\bibitem[Gururangan et~al\mbox{.}(2020)]%
        {dontstoppretrainingadapt.corr20}
\bibfield{author}{\bibinfo{person}{Suchin Gururangan}, \bibinfo{person}{Ana Marasović}, \bibinfo{person}{Swabha Swayamdipta}, \bibinfo{person}{Kyle Lo}, \bibinfo{person}{Iz Beltagy}, \bibinfo{person}{Doug Downey}, {and} \bibinfo{person}{Noah~A. Smith}.} \bibinfo{year}{2020}\natexlab{}.
\newblock \bibinfo{title}{Don't Stop Pretraining: Adapt Language Models to Domains and Tasks}.
\newblock
\showeprint[arxiv]{2004.10964}~[cs.CL]
\urldef\tempurl%
\url{https://arxiv.org/abs/2004.10964}
\showURL{%
\tempurl}


\bibitem[Hu et~al\mbox{.}(2021)]%
        {lora.corr21}
\bibfield{author}{\bibinfo{person}{Edward~J. Hu}, \bibinfo{person}{Yelong Shen}, \bibinfo{person}{Phillip Wallis}, \bibinfo{person}{Zeyuan Allen-Zhu}, \bibinfo{person}{Yuanzhi Li}, \bibinfo{person}{Shean Wang}, \bibinfo{person}{Lu Wang}, {and} \bibinfo{person}{Weizhu Chen}.} \bibinfo{year}{2021}\natexlab{}.
\newblock \bibinfo{title}{LoRA: Low-Rank Adaptation of Large Language Models}.
\newblock
\showeprint[arxiv]{2106.09685}~[cs.CL]
\urldef\tempurl%
\url{https://arxiv.org/abs/2106.09685}
\showURL{%
\tempurl}


\bibitem[Huang et~al\mbox{.}(2024)]%
        {Huang2024}
\bibfield{author}{\bibinfo{person}{Jun Huang}, \bibinfo{person}{Zhen Zhang}, \bibinfo{person}{Shuai Zheng}, \bibinfo{person}{Feng Qin}, {and} \bibinfo{person}{Yida Wang}.} \bibinfo{year}{2024}\natexlab{}.
\newblock \showarticletitle{DISTMM: Accelerating distributed multimodal model training}.
\newblock  (\bibinfo{year}{2024}).
\newblock
\urldef\tempurl%
\url{https://www.amazon.science/publications/distmm-accelerating-distributed-multimodal-model-training}
\showURL{%
\tempurl}


\bibitem[Huang et~al\mbox{.}(2019)]%
        {pipelineparallelism.corr19}
\bibfield{author}{\bibinfo{person}{Yanping Huang}, \bibinfo{person}{Youlong Cheng}, \bibinfo{person}{Ankur Bapna}, \bibinfo{person}{Orhan Firat}, \bibinfo{person}{Mia~Xu Chen}, \bibinfo{person}{Dehao Chen}, \bibinfo{person}{HyoukJoong Lee}, \bibinfo{person}{Jiquan Ngiam}, \bibinfo{person}{Quoc~V. Le}, \bibinfo{person}{Yonghui Wu}, {and} \bibinfo{person}{Zhifeng Chen}.} \bibinfo{year}{2019}\natexlab{}.
\newblock \bibinfo{title}{GPipe: Efficient Training of Giant Neural Networks using Pipeline Parallelism}.
\newblock
\showeprint[arxiv]{1811.06965}~[cs.CV]
\urldef\tempurl%
\url{https://arxiv.org/abs/1811.06965}
\showURL{%
\tempurl}


\bibitem[Inan et~al\mbox{.}(2023)]%
        {llamaguard.corr23}
\bibfield{author}{\bibinfo{person}{Hakan Inan}, \bibinfo{person}{Kartikeya Upasani}, \bibinfo{person}{Jianfeng Chi}, \bibinfo{person}{Rashi Rungta}, \bibinfo{person}{Krithika Iyer}, \bibinfo{person}{Yuning Mao}, \bibinfo{person}{Michael Tontchev}, \bibinfo{person}{Qing Hu}, \bibinfo{person}{Brian Fuller}, \bibinfo{person}{Davide Testuggine}, {and} \bibinfo{person}{Madian Khabsa}.} \bibinfo{year}{2023}\natexlab{}.
\newblock \bibinfo{title}{Llama Guard: LLM-based Input-Output Safeguard for Human-AI Conversations}.
\newblock
\showeprint[arxiv]{2312.06674}~[cs.CL]
\urldef\tempurl%
\url{https://arxiv.org/abs/2312.06674}
\showURL{%
\tempurl}


\bibitem[Ji et~al\mbox{.}(2024)]%
        {beavertails.nips24}
\bibfield{author}{\bibinfo{person}{Jiaming Ji}, \bibinfo{person}{Mickel Liu}, \bibinfo{person}{Josef Dai}, \bibinfo{person}{Xuehai Pan}, \bibinfo{person}{Chi Zhang}, \bibinfo{person}{Ce Bian}, \bibinfo{person}{Boyuan Chen}, \bibinfo{person}{Ruiyang Sun}, \bibinfo{person}{Yizhou Wang}, {and} \bibinfo{person}{Yaodong Yang}.} \bibinfo{year}{2024}\natexlab{}.
\newblock \showarticletitle{Beavertails: Towards improved safety alignment of llm via a human-preference dataset}.
\newblock \bibinfo{journal}{\emph{Advances in Neural Information Processing Systems}}  \bibinfo{volume}{36} (\bibinfo{year}{2024}).
\newblock


\bibitem[Jiang et~al\mbox{.}(2023)]%
        {mistral7b.corr23}
\bibfield{author}{\bibinfo{person}{Albert~Q. Jiang}, \bibinfo{person}{Alexandre Sablayrolles}, \bibinfo{person}{Arthur Mensch}, \bibinfo{person}{Chris Bamford}, \bibinfo{person}{Devendra~Singh Chaplot}, \bibinfo{person}{Diego de~las Casas}, \bibinfo{person}{Florian Bressand}, \bibinfo{person}{Gianna Lengyel}, \bibinfo{person}{Guillaume Lample}, \bibinfo{person}{Lucile Saulnier}, \bibinfo{person}{Lélio~Renard Lavaud}, \bibinfo{person}{Marie-Anne Lachaux}, \bibinfo{person}{Pierre Stock}, \bibinfo{person}{Teven~Le Scao}, \bibinfo{person}{Thibaut Lavril}, \bibinfo{person}{Thomas Wang}, \bibinfo{person}{Timothée Lacroix}, {and} \bibinfo{person}{William~El Sayed}.} \bibinfo{year}{2023}\natexlab{}.
\newblock \bibinfo{title}{Mistral 7B}.
\newblock
\showeprint[arxiv]{2310.06825}~[cs.CL]
\urldef\tempurl%
\url{https://arxiv.org/abs/2310.06825}
\showURL{%
\tempurl}


\bibitem[Jin et~al\mbox{.}(2024)]%
        {ragcache}
\bibfield{author}{\bibinfo{person}{Chao Jin}, \bibinfo{person}{Zili Zhang}, \bibinfo{person}{Xuanlin Jiang}, \bibinfo{person}{Fangyue Liu}, \bibinfo{person}{Xin Liu}, \bibinfo{person}{Xuanzhe Liu}, {and} \bibinfo{person}{Xin Jin}.} \bibinfo{year}{2024}\natexlab{}.
\newblock \bibinfo{title}{RAGCache: Efficient Knowledge Caching for Retrieval-Augmented Generation}.
\newblock
\showeprint[arxiv]{2404.12457}~[cs.DC]
\urldef\tempurl%
\url{https://arxiv.org/abs/2404.12457}
\showURL{%
\tempurl}


\bibitem[Kenton et~al\mbox{.}(2021)]%
        {alignmentlanguageagents.corr21}
\bibfield{author}{\bibinfo{person}{Zachary Kenton}, \bibinfo{person}{Tom Everitt}, \bibinfo{person}{Laura Weidinger}, \bibinfo{person}{Iason Gabriel}, \bibinfo{person}{Vladimir Mikulik}, {and} \bibinfo{person}{Geoffrey Irving}.} \bibinfo{year}{2021}\natexlab{}.
\newblock \bibinfo{title}{Alignment of Language Agents}.
\newblock
\showeprint[arxiv]{2103.14659}~[cs.AI]
\urldef\tempurl%
\url{https://arxiv.org/abs/2103.14659}
\showURL{%
\tempurl}


\bibitem[Khani et~al\mbox{.}(2023)]%
        {recl}
\bibfield{author}{\bibinfo{person}{Mehrdad Khani}, \bibinfo{person}{Ganesh Ananthanarayanan}, \bibinfo{person}{Kevin Hsieh}, \bibinfo{person}{Junchen Jiang}, \bibinfo{person}{Ravi Netravali}, \bibinfo{person}{Yuanchao Shu}, \bibinfo{person}{Mohammad Alizadeh}, {and} \bibinfo{person}{Victor Bahl}.} \bibinfo{year}{2023}\natexlab{}.
\newblock \showarticletitle{{RECL}: Responsive {Resource-Efficient} Continuous Learning for Video Analytics}. In \bibinfo{booktitle}{\emph{20th USENIX Symposium on Networked Systems Design and Implementation (NSDI 23)}}. \bibinfo{publisher}{USENIX Association}, \bibinfo{address}{Boston, MA}, \bibinfo{pages}{917--932}.
\newblock
\showISBNx{978-1-939133-33-5}
\urldef\tempurl%
\url{https://www.usenix.org/conference/nsdi23/presentation/khani}
\showURL{%
\tempurl}


\bibitem[Kim and Lee(2024)]%
        {adversarialdpo.corr24}
\bibfield{author}{\bibinfo{person}{San Kim} {and} \bibinfo{person}{Gary~Geunbae Lee}.} \bibinfo{year}{2024}\natexlab{}.
\newblock \bibinfo{title}{Adversarial DPO: Harnessing Harmful Data for Reducing Toxicity with Minimal Impact on Coherence and Evasiveness in Dialogue Agents}.
\newblock
\showeprint[arxiv]{2405.12900}~[cs.CL]
\urldef\tempurl%
\url{https://arxiv.org/abs/2405.12900}
\showURL{%
\tempurl}


\bibitem[Kopiczko et~al\mbox{.}(2024)]%
        {vera.iclr24}
\bibfield{author}{\bibinfo{person}{Dawid~Jan Kopiczko}, \bibinfo{person}{Tijmen Blankevoort}, {and} \bibinfo{person}{Yuki~M Asano}.} \bibinfo{year}{2024}\natexlab{}.
\newblock \showarticletitle{Ve{RA}: Vector-based Random Matrix Adaptation}. In \bibinfo{booktitle}{\emph{The Twelfth International Conference on Learning Representations}}.
\newblock
\urldef\tempurl%
\url{https://openreview.net/forum?id=NjNfLdxr3A}
\showURL{%
\tempurl}


\bibitem[{Kubernetes}(2025)]%
        {k8sautoscale}
\bibfield{author}{\bibinfo{person}{{Kubernetes}}.} \bibinfo{year}{2025}\natexlab{}.
\newblock \bibinfo{title}{Horizontal Pod Autoscale}.
\newblock \bibinfo{howpublished}{\url{https://kubernetes.io/docs/tasks/run-application/horizontal-pod-autoscale/}}.
\newblock
\newblock
\shownote{Kubernetes Documentation. Accessed: 2025-02-17}.


\bibitem[Kwon et~al\mbox{.}(2023)]%
        {vllm}
\bibfield{author}{\bibinfo{person}{Woosuk Kwon}, \bibinfo{person}{Zhuohan Li}, \bibinfo{person}{Siyuan Zhuang}, \bibinfo{person}{Ying Sheng}, \bibinfo{person}{Lianmin Zheng}, \bibinfo{person}{Cody~Hao Yu}, \bibinfo{person}{Joseph~E. Gonzalez}, \bibinfo{person}{Hao Zhang}, {and} \bibinfo{person}{Ion Stoica}.} \bibinfo{year}{2023}\natexlab{}.
\newblock \showarticletitle{Efficient Memory Management for Large Language Model Serving with PagedAttention}. In \bibinfo{booktitle}{\emph{Proceedings of the ACM SIGOPS 29th Symposium on Operating Systems Principles}}.
\newblock


\bibitem[Lee et~al\mbox{.}(2024b)]%
        {rlaif.corr24}
\bibfield{author}{\bibinfo{person}{Harrison Lee}, \bibinfo{person}{Samrat Phatale}, \bibinfo{person}{Hassan Mansoor}, \bibinfo{person}{Thomas Mesnard}, \bibinfo{person}{Johan Ferret}, \bibinfo{person}{Kellie Lu}, \bibinfo{person}{Colton Bishop}, \bibinfo{person}{Ethan Hall}, \bibinfo{person}{Victor Carbune}, \bibinfo{person}{Abhinav Rastogi}, {and} \bibinfo{person}{Sushant Prakash}.} \bibinfo{year}{2024}\natexlab{b}.
\newblock \bibinfo{title}{RLAIF vs. RLHF: Scaling Reinforcement Learning from Human Feedback with AI Feedback}.
\newblock
\showeprint[arxiv]{2309.00267}~[cs.CL]
\urldef\tempurl%
\url{https://arxiv.org/abs/2309.00267}
\showURL{%
\tempurl}


\bibitem[Lee et~al\mbox{.}(2024a)]%
        {infinigen}
\bibfield{author}{\bibinfo{person}{Wonbeom Lee}, \bibinfo{person}{Jungi Lee}, \bibinfo{person}{Junghwan Seo}, {and} \bibinfo{person}{Jaewoong Sim}.} \bibinfo{year}{2024}\natexlab{a}.
\newblock \bibinfo{title}{InfiniGen: Efficient Generative Inference of Large Language Models with Dynamic KV Cache Management}.
\newblock
\showeprint[arxiv]{2406.19707}~[cs.LG]
\urldef\tempurl%
\url{https://arxiv.org/abs/2406.19707}
\showURL{%
\tempurl}


\bibitem[Lefaudeux et~al\mbox{.}(2022)]%
        {xFormers2022}
\bibfield{author}{\bibinfo{person}{Benjamin Lefaudeux}, \bibinfo{person}{Francisco Massa}, \bibinfo{person}{Diana Liskovich}, \bibinfo{person}{Wenhan Xiong}, \bibinfo{person}{Vittorio Caggiano}, \bibinfo{person}{Sean Naren}, \bibinfo{person}{Min Xu}, \bibinfo{person}{Jieru Hu}, \bibinfo{person}{Marta Tintore}, \bibinfo{person}{Susan Zhang}, \bibinfo{person}{Patrick Labatut}, \bibinfo{person}{Daniel Haziza}, \bibinfo{person}{Luca Wehrstedt}, \bibinfo{person}{Jeremy Reizenstein}, {and} \bibinfo{person}{Grigory Sizov}.} \bibinfo{year}{2022}\natexlab{}.
\newblock \bibinfo{title}{xFormers: A modular and hackable Transformer modelling library}.
\newblock \bibinfo{howpublished}{\url{https://github.com/facebookresearch/xformers}}.
\newblock


\bibitem[Lester et~al\mbox{.}(2021)]%
        {prompttunning.emnlp21}
\bibfield{author}{\bibinfo{person}{Brian Lester}, \bibinfo{person}{Rami Al-Rfou}, {and} \bibinfo{person}{Noah Constant}.} \bibinfo{year}{2021}\natexlab{}.
\newblock \showarticletitle{The Power of Scale for Parameter-Efficient Prompt Tuning}. In \bibinfo{booktitle}{\emph{Proceedings of the 2021 Conference on Empirical Methods in Natural Language Processing}}, \bibfield{editor}{\bibinfo{person}{Marie-Francine Moens}, \bibinfo{person}{Xuanjing Huang}, \bibinfo{person}{Lucia Specia}, {and} \bibinfo{person}{Scott Wen-tau Yih}} (Eds.). \bibinfo{publisher}{Association for Computational Linguistics}, \bibinfo{address}{Online and Punta Cana, Dominican Republic}, \bibinfo{pages}{3045--3059}.
\newblock
\href{https://doi.org/10.18653/v1/2021.emnlp-main.243}{doi:\nolinkurl{10.18653/v1/2021.emnlp-main.243}}


\bibitem[Li et~al\mbox{.}(2024)]%
        {li2024demystifyingworkloadimbalanceslarge}
\bibfield{author}{\bibinfo{person}{Haoyang Li}, \bibinfo{person}{Fangcheng Fu}, \bibinfo{person}{Sheng Lin}, \bibinfo{person}{Hao Ge}, \bibinfo{person}{Xuanyu Wang}, \bibinfo{person}{Jiawen Niu}, \bibinfo{person}{Jie Jiang}, {and} \bibinfo{person}{Bin Cui}.} \bibinfo{year}{2024}\natexlab{}.
\newblock \bibinfo{title}{Demystifying Workload Imbalances in Large Transformer Model Training over Variable-length Sequences}.
\newblock
\showeprint[arxiv]{2412.07894}~[cs.DC]
\urldef\tempurl%
\url{https://arxiv.org/abs/2412.07894}
\showURL{%
\tempurl}


\bibitem[Li and Liang(2021)]%
        {prefixtunning.acl21}
\bibfield{author}{\bibinfo{person}{Xiang~Lisa Li} {and} \bibinfo{person}{Percy Liang}.} \bibinfo{year}{2021}\natexlab{}.
\newblock \showarticletitle{Prefix-Tuning: Optimizing Continuous Prompts for Generation}. In \bibinfo{booktitle}{\emph{Proceedings of the 59th Annual Meeting of the Association for Computational Linguistics and the 11th International Joint Conference on Natural Language Processing (Volume 1: Long Papers)}}, \bibfield{editor}{\bibinfo{person}{Chengqing Zong}, \bibinfo{person}{Fei Xia}, \bibinfo{person}{Wenjie Li}, {and} \bibinfo{person}{Roberto Navigli}} (Eds.). \bibinfo{publisher}{Association for Computational Linguistics}, \bibinfo{address}{Online}, \bibinfo{pages}{4582--4597}.
\newblock
\href{https://doi.org/10.18653/v1/2021.acl-long.353}{doi:\nolinkurl{10.18653/v1/2021.acl-long.353}}


\bibitem[Lian et~al\mbox{.}(2024)]%
        {lian2024universalcheckpointingefficientflexible}
\bibfield{author}{\bibinfo{person}{Xinyu Lian}, \bibinfo{person}{Sam~Ade Jacobs}, \bibinfo{person}{Lev Kurilenko}, \bibinfo{person}{Masahiro Tanaka}, \bibinfo{person}{Stas Bekman}, \bibinfo{person}{Olatunji Ruwase}, {and} \bibinfo{person}{Minjia Zhang}.} \bibinfo{year}{2024}\natexlab{}.
\newblock \bibinfo{title}{Universal Checkpointing: Efficient and Flexible Checkpointing for Large Scale Distributed Training}.
\newblock
\showeprint[arxiv]{2406.18820}~[cs.DC]
\urldef\tempurl%
\url{https://arxiv.org/abs/2406.18820}
\showURL{%
\tempurl}


\bibitem[Lin et~al\mbox{.}(2024a)]%
        {parrot}
\bibfield{author}{\bibinfo{person}{Chaofan Lin}, \bibinfo{person}{Zhenhua Han}, \bibinfo{person}{Chengruidong Zhang}, \bibinfo{person}{Yuqing Yang}, \bibinfo{person}{Fan Yang}, \bibinfo{person}{Chen Chen}, {and} \bibinfo{person}{Lili Qiu}.} \bibinfo{year}{2024}\natexlab{a}.
\newblock \bibinfo{title}{Parrot: Efficient Serving of LLM-based Applications with Semantic Variable}.
\newblock
\showeprint[arxiv]{2405.19888}~[cs.LG]
\urldef\tempurl%
\url{https://arxiv.org/abs/2405.19888}
\showURL{%
\tempurl}


\bibitem[Lin et~al\mbox{.}(2024b)]%
        {qserve.corr24}
\bibfield{author}{\bibinfo{person}{Yujun Lin}, \bibinfo{person}{Haotian Tang}, \bibinfo{person}{Shang Yang}, \bibinfo{person}{Zhekai Zhang}, \bibinfo{person}{Guangxuan Xiao}, \bibinfo{person}{Chuang Gan}, {and} \bibinfo{person}{Song Han}.} \bibinfo{year}{2024}\natexlab{b}.
\newblock \bibinfo{title}{QServe: W4A8KV4 Quantization and System Co-design for Efficient LLM Serving}.
\newblock
\showeprint[arxiv]{2405.04532}~[cs.CL]
\urldef\tempurl%
\url{https://arxiv.org/abs/2405.04532}
\showURL{%
\tempurl}


\bibitem[Liu et~al\mbox{.}(2023a)]%
        {ptuning.corr23}
\bibfield{author}{\bibinfo{person}{Xiao Liu}, \bibinfo{person}{Yanan Zheng}, \bibinfo{person}{Zhengxiao Du}, \bibinfo{person}{Ming Ding}, \bibinfo{person}{Yujie Qian}, \bibinfo{person}{Zhilin Yang}, {and} \bibinfo{person}{Jie Tang}.} \bibinfo{year}{2023}\natexlab{a}.
\newblock \bibinfo{title}{GPT Understands, Too}.
\newblock
\showeprint[arxiv]{2103.10385}~[cs.CL]
\urldef\tempurl%
\url{https://arxiv.org/abs/2103.10385}
\showURL{%
\tempurl}


\bibitem[Liu et~al\mbox{.}(2023b)]%
        {multitaskprompttuning.corr23}
\bibfield{author}{\bibinfo{person}{Xiao Liu}, \bibinfo{person}{Yanan Zheng}, \bibinfo{person}{Zhengxiao Du}, \bibinfo{person}{Ming Ding}, \bibinfo{person}{Yujie Qian}, \bibinfo{person}{Zhilin Yang}, {and} \bibinfo{person}{Jie Tang}.} \bibinfo{year}{2023}\natexlab{b}.
\newblock \bibinfo{title}{GPT Understands, Too}.
\newblock
\showeprint[arxiv]{2103.10385}~[cs.CL]
\urldef\tempurl%
\url{https://arxiv.org/abs/2103.10385}
\showURL{%
\tempurl}


\bibitem[Liu et~al\mbox{.}(2024)]%
        {cachegen}
\bibfield{author}{\bibinfo{person}{Yuhan Liu}, \bibinfo{person}{Hanchen Li}, \bibinfo{person}{Yihua Cheng}, \bibinfo{person}{Siddhant Ray}, \bibinfo{person}{Yuyang Huang}, \bibinfo{person}{Qizheng Zhang}, \bibinfo{person}{Kuntai Du}, \bibinfo{person}{Jiayi Yao}, \bibinfo{person}{Shan Lu}, \bibinfo{person}{Ganesh Ananthanarayanan}, \bibinfo{person}{Michael Maire}, \bibinfo{person}{Henry Hoffmann}, \bibinfo{person}{Ari Holtzman}, {and} \bibinfo{person}{Junchen Jiang}.} \bibinfo{year}{2024}\natexlab{}.
\newblock \bibinfo{title}{CacheGen: KV Cache Compression and Streaming for Fast Large Language Model Serving}.
\newblock
\showeprint[arxiv]{2310.07240}~[cs.NI]
\urldef\tempurl%
\url{https://arxiv.org/abs/2310.07240}
\showURL{%
\tempurl}


\bibitem[Love(2025)]%
        {taskset}
\bibfield{author}{\bibinfo{person}{Robert~M. Love}.} \bibinfo{year}{2025}\natexlab{}.
\newblock \bibinfo{title}{taskset(1) -- set or retrieve a process's CPU affinity}.
\newblock \bibinfo{howpublished}{\url{https://man7.org/linux/man-pages/man1/taskset.1.html}}.
\newblock
\newblock
\shownote{Linux manual page. Accessed: 2025-02-17}.


\bibitem[Miao et~al\mbox{.}(2024a)]%
        {flexllm}
\bibfield{author}{\bibinfo{person}{Xupeng Miao}, \bibinfo{person}{Gabriele Oliaro}, \bibinfo{person}{Xinhao Cheng}, \bibinfo{person}{Mengdi Wu}, \bibinfo{person}{Colin Unger}, {and} \bibinfo{person}{Zhihao Jia}.} \bibinfo{year}{2024}\natexlab{a}.
\newblock \bibinfo{title}{FlexLLM: A System for Co-Serving Large Language Model Inference and Parameter-Efficient Finetuning}.
\newblock
\showeprint[arxiv]{2402.18789}~[cs.DC]
\urldef\tempurl%
\url{https://arxiv.org/abs/2402.18789}
\showURL{%
\tempurl}


\bibitem[Miao et~al\mbox{.}(2024b)]%
        {specinfer}
\bibfield{author}{\bibinfo{person}{Xupeng Miao}, \bibinfo{person}{Gabriele Oliaro}, \bibinfo{person}{Zhihao Zhang}, \bibinfo{person}{Xinhao Cheng}, \bibinfo{person}{Zeyu Wang}, \bibinfo{person}{Zhengxin Zhang}, \bibinfo{person}{Rae Ying~Yee Wong}, \bibinfo{person}{Alan Zhu}, \bibinfo{person}{Lijie Yang}, \bibinfo{person}{Xiaoxiang Shi}, \bibinfo{person}{Chunan Shi}, \bibinfo{person}{Zhuoming Chen}, \bibinfo{person}{Daiyaan Arfeen}, \bibinfo{person}{Reyna Abhyankar}, {and} \bibinfo{person}{Zhihao Jia}.} \bibinfo{year}{2024}\natexlab{b}.
\newblock \showarticletitle{SpecInfer: Accelerating Large Language Model Serving with Tree-based Speculative Inference and Verification}. In \bibinfo{booktitle}{\emph{Proceedings of the 29th ACM International Conference on Architectural Support for Programming Languages and Operating Systems, Volume 3}} \emph{(\bibinfo{series}{ASPLOS ’24})}. \bibinfo{publisher}{ACM}, \bibinfo{pages}{932–949}.
\newblock
\href{https://doi.org/10.1145/3620666.3651335}{doi:\nolinkurl{10.1145/3620666.3651335}}


\bibitem[Microsoft(nd)]%
        {microsoft_mscclpp}
\bibfield{author}{\bibinfo{person}{Microsoft}.} \bibinfo{year}{n.d.}\natexlab{}.
\newblock \bibinfo{title}{mscclpp}.
\newblock \bibinfo{howpublished}{\url{https://github.com/microsoft/mscclpp}}.
\newblock
\newblock
\shownote{GitHub repository. Accessed: March 1, 2025}.


\bibitem[Morimura et~al\mbox{.}(2024)]%
        {filtereddpo.corr24}
\bibfield{author}{\bibinfo{person}{Tetsuro Morimura}, \bibinfo{person}{Mitsuki Sakamoto}, \bibinfo{person}{Yuu Jinnai}, \bibinfo{person}{Kenshi Abe}, {and} \bibinfo{person}{Kaito Ariu}.} \bibinfo{year}{2024}\natexlab{}.
\newblock \bibinfo{title}{Filtered Direct Preference Optimization}.
\newblock
\showeprint[arxiv]{2404.13846}~[cs.LG]
\urldef\tempurl%
\url{https://arxiv.org/abs/2404.13846}
\showURL{%
\tempurl}


\bibitem[{NVIDIA}(2025)]%
        {a100gpu}
\bibfield{author}{\bibinfo{person}{{NVIDIA}}.} \bibinfo{year}{2025}\natexlab{}.
\newblock \bibinfo{title}{NVIDIA A100 Data Center GPU}.
\newblock \bibinfo{howpublished}{\url{https://www.nvidia.com/en-us/data-center/a100/}}.
\newblock
\newblock
\shownote{Accessed: 2025-02-17}.


\bibitem[Ousidhoum et~al\mbox{.}(2021)]%
        {probingtoxiccontent.acl21}
\bibfield{author}{\bibinfo{person}{Nedjma Ousidhoum}, \bibinfo{person}{Xinran Zhao}, \bibinfo{person}{Tianqing Fang}, \bibinfo{person}{Yangqiu Song}, {and} \bibinfo{person}{Dit-Yan Yeung}.} \bibinfo{year}{2021}\natexlab{}.
\newblock \showarticletitle{Probing toxic content in large pre-trained language models}. In \bibinfo{booktitle}{\emph{Proceedings of the 59th Annual Meeting of the Association for Computational Linguistics and the 11th International Joint Conference on Natural Language Processing (Volume 1: Long Papers)}}. \bibinfo{pages}{4262--4274}.
\newblock


\bibitem[Patel et~al\mbox{.}(2023)]%
        {splitwise}
\bibfield{author}{\bibinfo{person}{Pratyush Patel}, \bibinfo{person}{Esha Choukse}, \bibinfo{person}{Chaojie Zhang}, \bibinfo{person}{{ \'I}{\~n}igo Goiri}, \bibinfo{person}{Aashaka Shah}, \bibinfo{person}{Saeed Maleki}, {and} \bibinfo{person}{Ricardo Bianchini}.} \bibinfo{year}{2023}\natexlab{}.
\newblock \showarticletitle{Splitwise: Efficient generative llm inference using phase splitting}.
\newblock \bibinfo{journal}{\emph{arXiv preprint arXiv:2311.18677}} (\bibinfo{year}{2023}).
\newblock


\bibitem[Peng et~al\mbox{.}(2023)]%
        {fp8training.corr23}
\bibfield{author}{\bibinfo{person}{Houwen Peng}, \bibinfo{person}{Kan Wu}, \bibinfo{person}{Yixuan Wei}, \bibinfo{person}{Guoshuai Zhao}, \bibinfo{person}{Yuxiang Yang}, \bibinfo{person}{Ze Liu}, \bibinfo{person}{Yifan Xiong}, \bibinfo{person}{Ziyue Yang}, \bibinfo{person}{Bolin Ni}, \bibinfo{person}{Jingcheng Hu}, \bibinfo{person}{Ruihang Li}, \bibinfo{person}{Miaosen Zhang}, \bibinfo{person}{Chen Li}, \bibinfo{person}{Jia Ning}, \bibinfo{person}{Ruizhe Wang}, \bibinfo{person}{Zheng Zhang}, \bibinfo{person}{Shuguang Liu}, \bibinfo{person}{Joe Chau}, \bibinfo{person}{Han Hu}, {and} \bibinfo{person}{Peng Cheng}.} \bibinfo{year}{2023}\natexlab{}.
\newblock \bibinfo{title}{FP8-LM: Training FP8 Large Language Models}.
\newblock
\showeprint[arxiv]{2310.18313}~[cs.LG]
\urldef\tempurl%
\url{https://arxiv.org/abs/2310.18313}
\showURL{%
\tempurl}


\bibitem[{PyTorch Contributors}(2025a)]%
        {pytorch_cuda_memory_management}
\bibfield{author}{\bibinfo{person}{{PyTorch Contributors}}.} \bibinfo{year}{2025}\natexlab{a}.
\newblock \bibinfo{title}{CUDA Memory Management}.
\newblock \bibinfo{howpublished}{\url{https://pytorch.org/docs/stable/notes/cuda.html\#cuda-memory-management}}.
\newblock
\newblock
\shownote{PyTorch Documentation, Version 2.5. Accessed: 2025-02-14}.


\bibitem[{PyTorch Contributors}(2025b)]%
        {pytorch_cuda_semantics}
\bibfield{author}{\bibinfo{person}{{PyTorch Contributors}}.} \bibinfo{year}{2025}\natexlab{b}.
\newblock \bibinfo{title}{CUDA Semantics}.
\newblock \bibinfo{howpublished}{\url{https://pytorch.org/docs/stable/notes/cuda.html}}.
\newblock
\newblock
\shownote{PyTorch 2.6 Documentation. Accessed: 2025-02-14}.


\bibitem[{PyTorch Tutorials}(2025)]%
        {pytorch_autograd_saved_tensors_hooks}
\bibfield{author}{\bibinfo{person}{{PyTorch Tutorials}}.} \bibinfo{year}{2025}\natexlab{}.
\newblock \bibinfo{title}{Autograd: Saved Tensors Hooks Tutorial}.
\newblock \bibinfo{howpublished}{\url{https://pytorch.org/tutorials/intermediate/autograd_saved_tensors_hooks_tutorial.html\#saved-tensors-hooks}}.
\newblock
\newblock
\shownote{PyTorch Tutorials. Accessed: 2025-02-14}.


\bibitem[Qi et~al\mbox{.}(2024a)]%
        {ofsdpo.corr24}
\bibfield{author}{\bibinfo{person}{Biqing Qi}, \bibinfo{person}{Pengfei Li}, \bibinfo{person}{Fangyuan Li}, \bibinfo{person}{Junqi Gao}, \bibinfo{person}{Kaiyan Zhang}, {and} \bibinfo{person}{Bowen Zhou}.} \bibinfo{year}{2024}\natexlab{a}.
\newblock \bibinfo{title}{Online DPO: Online Direct Preference Optimization with Fast-Slow Chasing}.
\newblock
\showeprint[arxiv]{2406.05534}~[cs.AI]
\urldef\tempurl%
\url{https://arxiv.org/abs/2406.05534}
\showURL{%
\tempurl}


\bibitem[Qi et~al\mbox{.}(2024b)]%
        {dpo.corr24}
\bibfield{author}{\bibinfo{person}{Biqing Qi}, \bibinfo{person}{Pengfei Li}, \bibinfo{person}{Fangyuan Li}, \bibinfo{person}{Junqi Gao}, \bibinfo{person}{Kaiyan Zhang}, {and} \bibinfo{person}{Bowen Zhou}.} \bibinfo{year}{2024}\natexlab{b}.
\newblock \bibinfo{title}{Online DPO: Online Direct Preference Optimization with Fast-Slow Chasing}.
\newblock
\showeprint[arxiv]{2406.05534}~[cs.AI]
\urldef\tempurl%
\url{https://arxiv.org/abs/2406.05534}
\showURL{%
\tempurl}


\bibitem[Qiao et~al\mbox{.}(2024)]%
        {conserve}
\bibfield{author}{\bibinfo{person}{Yifan Qiao}, \bibinfo{person}{Shu Anzai}, \bibinfo{person}{Shan Yu}, \bibinfo{person}{Haoran Ma}, \bibinfo{person}{Yang Wang}, \bibinfo{person}{Miryung Kim}, {and} \bibinfo{person}{Harry Xu}.} \bibinfo{year}{2024}\natexlab{}.
\newblock \bibinfo{title}{ConServe: Harvesting GPUs for Low-Latency and High-Throughput Large Language Model Serving}.
\newblock
\showeprint[arxiv]{2410.01228}~[cs.DC]
\urldef\tempurl%
\url{https://arxiv.org/abs/2410.01228}
\showURL{%
\tempurl}


\bibitem[Qin et~al\mbox{.}(2025)]%
        {mooncake.fast25}
\bibfield{author}{\bibinfo{person}{Ruoyu Qin}, \bibinfo{person}{Zheming Li}, \bibinfo{person}{Weiran He}, \bibinfo{person}{Jialei Cui}, \bibinfo{person}{Feng Ren}, \bibinfo{person}{Mingxing Zhang}, \bibinfo{person}{Yongwei Wu}, \bibinfo{person}{Weimin Zheng}, {and} \bibinfo{person}{Xinran Xu}.} \bibinfo{year}{2025}\natexlab{}.
\newblock \showarticletitle{Mooncake: Trading More Storage for Less Computation {\textemdash} A {KVCache-centric} Architecture for Serving {LLM} Chatbot}. In \bibinfo{booktitle}{\emph{23rd USENIX Conference on File and Storage Technologies (FAST 25)}}. \bibinfo{publisher}{USENIX Association}, \bibinfo{address}{Santa Clara, CA}, \bibinfo{pages}{155--170}.
\newblock
\showISBNx{978-1-939133-45-8}
\urldef\tempurl%
\url{https://www.usenix.org/conference/fast25/presentation/qin}
\showURL{%
\tempurl}


\bibitem[Qiu et~al\mbox{.}(2023)]%
        {oft.nips23}
\bibfield{author}{\bibinfo{person}{Zeju Qiu}, \bibinfo{person}{Weiyang Liu}, \bibinfo{person}{Haiwen Feng}, \bibinfo{person}{Yuxuan Xue}, \bibinfo{person}{Yao Feng}, \bibinfo{person}{Zhen Liu}, \bibinfo{person}{Dan Zhang}, \bibinfo{person}{Adrian Weller}, {and} \bibinfo{person}{Bernhard Sch{\"o}lkopf}.} \bibinfo{year}{2023}\natexlab{}.
\newblock \showarticletitle{Controlling Text-to-Image Diffusion by Orthogonal Finetuning}. In \bibinfo{booktitle}{\emph{NeurIPS}}.
\newblock


\bibitem[Ramachandran et~al\mbox{.}(2024)]%
        {microscopiq.corr24}
\bibfield{author}{\bibinfo{person}{Akshat Ramachandran}, \bibinfo{person}{Souvik Kundu}, {and} \bibinfo{person}{Tushar Krishna}.} \bibinfo{year}{2024}\natexlab{}.
\newblock \bibinfo{title}{MicroScopiQ: Accelerating Foundational Models through Outlier-Aware Microscaling Quantization}.
\newblock
\showeprint[arxiv]{2411.05282}~[cs.AR]
\urldef\tempurl%
\url{https://arxiv.org/abs/2411.05282}
\showURL{%
\tempurl}


\bibitem[Schulman et~al\mbox{.}(2017)]%
        {ppo.corr17}
\bibfield{author}{\bibinfo{person}{John Schulman}, \bibinfo{person}{Filip Wolski}, \bibinfo{person}{Prafulla Dhariwal}, \bibinfo{person}{Alec Radford}, {and} \bibinfo{person}{Oleg Klimov}.} \bibinfo{year}{2017}\natexlab{}.
\newblock \bibinfo{title}{Proximal Policy Optimization Algorithms}.
\newblock
\showeprint[arxiv]{1707.06347}~[cs.LG]
\urldef\tempurl%
\url{https://arxiv.org/abs/1707.06347}
\showURL{%
\tempurl}


\bibitem[Shaikh et~al\mbox{.}(2024)]%
        {showdonttellaligning.corr24}
\bibfield{author}{\bibinfo{person}{Omar Shaikh}, \bibinfo{person}{Michelle Lam}, \bibinfo{person}{Joey Hejna}, \bibinfo{person}{Yijia Shao}, \bibinfo{person}{Michael Bernstein}, {and} \bibinfo{person}{Diyi Yang}.} \bibinfo{year}{2024}\natexlab{}.
\newblock \bibinfo{title}{Show, Don't Tell: Aligning Language Models with Demonstrated Feedback}.
\newblock
\showeprint[arxiv]{2406.00888}~[cs.CL]
\urldef\tempurl%
\url{https://arxiv.org/abs/2406.00888}
\showURL{%
\tempurl}


\bibitem[Song et~al\mbox{.}(2023)]%
        {conpet.corr23}
\bibfield{author}{\bibinfo{person}{Chenyang Song}, \bibinfo{person}{Xu Han}, \bibinfo{person}{Zheni Zeng}, \bibinfo{person}{Kuai Li}, \bibinfo{person}{Chen Chen}, \bibinfo{person}{Zhiyuan Liu}, \bibinfo{person}{Maosong Sun}, {and} \bibinfo{person}{Tao Yang}.} \bibinfo{year}{2023}\natexlab{}.
\newblock \bibinfo{title}{ConPET: Continual Parameter-Efficient Tuning for Large Language Models}.
\newblock
\showeprint[arxiv]{2309.14763}~[cs.CL]
\urldef\tempurl%
\url{https://arxiv.org/abs/2309.14763}
\showURL{%
\tempurl}


\bibitem[Srivatsa et~al\mbox{.}(2024)]%
        {preble}
\bibfield{author}{\bibinfo{person}{Vikranth Srivatsa}, \bibinfo{person}{Zijian He}, \bibinfo{person}{Reyna Abhyankar}, \bibinfo{person}{Dongming Li}, {and} \bibinfo{person}{Yiying Zhang}.} \bibinfo{year}{2024}\natexlab{}.
\newblock \bibinfo{title}{Preble: Efficient Distributed Prompt Scheduling for LLM Serving}.
\newblock
\showeprint[arxiv]{2407.00023}~[cs.DC]
\urldef\tempurl%
\url{https://arxiv.org/abs/2407.00023}
\showURL{%
\tempurl}


\bibitem[Stiennon et~al\mbox{.}(2022)]%
        {learningsummarizehumanfeedback.corr22}
\bibfield{author}{\bibinfo{person}{Nisan Stiennon}, \bibinfo{person}{Long Ouyang}, \bibinfo{person}{Jeff Wu}, \bibinfo{person}{Daniel~M. Ziegler}, \bibinfo{person}{Ryan Lowe}, \bibinfo{person}{Chelsea Voss}, \bibinfo{person}{Alec Radford}, \bibinfo{person}{Dario Amodei}, {and} \bibinfo{person}{Paul Christiano}.} \bibinfo{year}{2022}\natexlab{}.
\newblock \bibinfo{title}{Learning to summarize from human feedback}.
\newblock
\showeprint[arxiv]{2009.01325}~[cs.CL]
\urldef\tempurl%
\url{https://arxiv.org/abs/2009.01325}
\showURL{%
\tempurl}


\bibitem[Sun et~al\mbox{.}(2019)]%
        {hybridfp8training.nips19}
\bibfield{author}{\bibinfo{person}{Xiao Sun}, \bibinfo{person}{Jungwook Choi}, \bibinfo{person}{Chia-Yu Chen}, \bibinfo{person}{Naigang Wang}, \bibinfo{person}{Swagath Venkataramani}, \bibinfo{person}{Vijayalakshmi~Viji Srinivasan}, \bibinfo{person}{Xiaodong Cui}, \bibinfo{person}{Wei Zhang}, {and} \bibinfo{person}{Kailash Gopalakrishnan}.} \bibinfo{year}{2019}\natexlab{}.
\newblock \showarticletitle{Hybrid 8-bit floating point (HFP8) training and inference for deep neural networks}.
\newblock \bibinfo{journal}{\emph{Advances in neural information processing systems}}  \bibinfo{volume}{32} (\bibinfo{year}{2019}).
\newblock


\bibitem[Sun et~al\mbox{.}(2020)]%
        {ernie.aaai20}
\bibfield{author}{\bibinfo{person}{Yu Sun}, \bibinfo{person}{Shuohuan Wang}, \bibinfo{person}{Yukun Li}, \bibinfo{person}{Shikun Feng}, \bibinfo{person}{Hao Tian}, \bibinfo{person}{Hua Wu}, {and} \bibinfo{person}{Haifeng Wang}.} \bibinfo{year}{2020}\natexlab{}.
\newblock \showarticletitle{Ernie 2.0: A continual pre-training framework for language understanding}. In \bibinfo{booktitle}{\emph{Proceedings of the AAAI conference on artificial intelligence}}, Vol.~\bibinfo{volume}{34}. \bibinfo{pages}{8968--8975}.
\newblock


\bibitem[{TechTarget Editorial}(2023)]%
        {techtarget2023llm}
\bibfield{author}{\bibinfo{person}{{TechTarget Editorial}}.} \bibinfo{year}{2023}\natexlab{}.
\newblock \bibinfo{booktitle}{\emph{Top 10 Large Language Models (LLMs) in 2024}}.
\newblock
\urldef\tempurl%
\url{https://www.techtarget.com}
\showURL{%
\tempurl}


\bibitem[Topics(2024)]%
        {exploding2024llm}
\bibfield{author}{\bibinfo{person}{Exploding Topics}.} \bibinfo{year}{2024}\natexlab{}.
\newblock \bibinfo{booktitle}{\emph{Large Language Model (LLM) Market Growth Analysis}}.
\newblock
\urldef\tempurl%
\url{https://explodingtopics.com}
\showURL{%
\tempurl}


\bibitem[Touvron et~al\mbox{.}(2023)]%
        {llamamodel.corr23}
\bibfield{author}{\bibinfo{person}{Hugo Touvron}, \bibinfo{person}{Thibaut Lavril}, \bibinfo{person}{Gautier Izacard}, \bibinfo{person}{Xavier Martinet}, \bibinfo{person}{Marie-Anne Lachaux}, \bibinfo{person}{Timothée Lacroix}, \bibinfo{person}{Baptiste Rozière}, \bibinfo{person}{Naman Goyal}, \bibinfo{person}{Eric Hambro}, \bibinfo{person}{Faisal Azhar}, \bibinfo{person}{Aurelien Rodriguez}, \bibinfo{person}{Armand Joulin}, \bibinfo{person}{Edouard Grave}, {and} \bibinfo{person}{Guillaume Lample}.} \bibinfo{year}{2023}\natexlab{}.
\newblock \bibinfo{title}{LLaMA: Open and Efficient Foundation Language Models}.
\newblock
\showeprint[arxiv]{2302.13971}~[cs.CL]
\urldef\tempurl%
\url{https://arxiv.org/abs/2302.13971}
\showURL{%
\tempurl}


\bibitem[Vaswani et~al\mbox{.}(2017)]%
        {transformers}
\bibfield{author}{\bibinfo{person}{Ashish Vaswani}, \bibinfo{person}{Noam Shazeer}, \bibinfo{person}{Niki Parmar}, \bibinfo{person}{Jakob Uszkoreit}, \bibinfo{person}{Llion Jones}, \bibinfo{person}{Aidan~N. Gomez}, \bibinfo{person}{Lukasz Kaiser}, {and} \bibinfo{person}{Illia Polosukhin}.} \bibinfo{year}{2017}\natexlab{}.
\newblock \showarticletitle{Attention is All You Need}.
\newblock
\urldef\tempurl%
\url{https://arxiv.org/pdf/1706.03762.pdf}
\showURL{%
\tempurl}


\bibitem[Wan et~al\mbox{.}(2024)]%
        {wan2024bytecheckpointunifiedcheckpointinglarge}
\bibfield{author}{\bibinfo{person}{Borui Wan}, \bibinfo{person}{Mingji Han}, \bibinfo{person}{Yiyao Sheng}, \bibinfo{person}{Yanghua Peng}, \bibinfo{person}{Haibin Lin}, \bibinfo{person}{Mofan Zhang}, \bibinfo{person}{Zhichao Lai}, \bibinfo{person}{Menghan Yu}, \bibinfo{person}{Junda Zhang}, \bibinfo{person}{Zuquan Song}, \bibinfo{person}{Xin Liu}, {and} \bibinfo{person}{Chuan Wu}.} \bibinfo{year}{2024}\natexlab{}.
\newblock \bibinfo{title}{ByteCheckpoint: A Unified Checkpointing System for Large Foundation Model Development}.
\newblock
\showeprint[arxiv]{2407.20143}~[cs.AI]
\urldef\tempurl%
\url{https://arxiv.org/abs/2407.20143}
\showURL{%
\tempurl}


\bibitem[Wang et~al\mbox{.}(2023b)]%
        {reversekl.iclr23}
\bibfield{author}{\bibinfo{person}{Chaoqi Wang}, \bibinfo{person}{Yibo Jiang}, \bibinfo{person}{Chenghao Yang}, \bibinfo{person}{Han Liu}, {and} \bibinfo{person}{Yuxin Chen}.} \bibinfo{year}{2023}\natexlab{b}.
\newblock \bibinfo{title}{Beyond Reverse KL: Generalizing Direct Preference Optimization with Diverse Divergence Constraints}.
\newblock
\showeprint[arxiv]{2309.16240}~[cs.LG]
\urldef\tempurl%
\url{https://arxiv.org/abs/2309.16240}
\showURL{%
\tempurl}


\bibitem[Wang et~al\mbox{.}(2025)]%
        {fp4training.corr25}
\bibfield{author}{\bibinfo{person}{Ruizhe Wang}, \bibinfo{person}{Yeyun Gong}, \bibinfo{person}{Xiao Liu}, \bibinfo{person}{Guoshuai Zhao}, \bibinfo{person}{Ziyue Yang}, \bibinfo{person}{Baining Guo}, \bibinfo{person}{Zhengjun Zha}, {and} \bibinfo{person}{Peng Cheng}.} \bibinfo{year}{2025}\natexlab{}.
\newblock \bibinfo{title}{Optimizing Large Language Model Training Using FP4 Quantization}.
\newblock
\showeprint[arxiv]{2501.17116}~[cs.LG]
\urldef\tempurl%
\url{https://arxiv.org/abs/2501.17116}
\showURL{%
\tempurl}


\bibitem[Wang et~al\mbox{.}(2023a)]%
        {recover}
\bibfield{author}{\bibinfo{person}{Zhuang Wang}, \bibinfo{person}{Zhen Jia}, \bibinfo{person}{Shuai Zheng}, \bibinfo{person}{Zhen Zhang}, \bibinfo{person}{Xinwei Fu}, \bibinfo{person}{T.~S.~Eugene Ng}, {and} \bibinfo{person}{Yida Wang}.} \bibinfo{year}{2023}\natexlab{a}.
\newblock \showarticletitle{GEMINI: Fast Failure Recovery in Distributed Training with In-Memory Checkpoints}. In \bibinfo{booktitle}{\emph{Proceedings of the 29th Symposium on Operating Systems Principles}} (Koblenz, Germany) \emph{(\bibinfo{series}{SOSP '23})}. \bibinfo{publisher}{Association for Computing Machinery}, \bibinfo{address}{New York, NY, USA}, \bibinfo{pages}{364–381}.
\newblock
\showISBNx{9798400702297}
\href{https://doi.org/10.1145/3600006.3613145}{doi:\nolinkurl{10.1145/3600006.3613145}}


\bibitem[Wolf et~al\mbox{.}(2020)]%
        {transformers.emnlp20}
\bibfield{author}{\bibinfo{person}{Thomas Wolf}, \bibinfo{person}{Lysandre Debut}, \bibinfo{person}{Victor Sanh}, \bibinfo{person}{Julien Chaumond}, \bibinfo{person}{Clement Delangue}, \bibinfo{person}{Anthony Moi}, \bibinfo{person}{Pierric Cistac}, \bibinfo{person}{Tim Rault}, \bibinfo{person}{Rémi Louf}, \bibinfo{person}{Morgan Funtowicz}, \bibinfo{person}{Joe Davison}, \bibinfo{person}{Sam Shleifer}, \bibinfo{person}{Patrick von Platen}, \bibinfo{person}{Clara Ma}, \bibinfo{person}{Yacine Jernite}, \bibinfo{person}{Julien Plu}, \bibinfo{person}{Canwen Xu}, \bibinfo{person}{Teven~Le Scao}, \bibinfo{person}{Sylvain Gugger}, \bibinfo{person}{Mariama Drame}, \bibinfo{person}{Quentin Lhoest}, {and} \bibinfo{person}{Alexander~M. Rush}.} \bibinfo{year}{2020}\natexlab{}.
\newblock \showarticletitle{Transformers: State-of-the-Art Natural Language Processing}. In \bibinfo{booktitle}{\emph{Proceedings of the 2020 Conference on Empirical Methods in Natural Language Processing: System Demonstrations}}. \bibinfo{publisher}{Association for Computational Linguistics}, \bibinfo{address}{Online}, \bibinfo{pages}{38--45}.
\newblock
\urldef\tempurl%
\url{https://www.aclweb.org/anthology/2020.emnlp-demos.6}
\showURL{%
\tempurl}


\bibitem[Xiang et~al\mbox{.}(2025)]%
        {metacot}
\bibfield{author}{\bibinfo{person}{Violet Xiang}, \bibinfo{person}{Charlie Snell}, \bibinfo{person}{Kanishk Gandhi}, \bibinfo{person}{Alon Albalak}, \bibinfo{person}{Anikait Singh}, \bibinfo{person}{Chase Blagden}, \bibinfo{person}{Duy Phung}, \bibinfo{person}{Rafael Rafailov}, \bibinfo{person}{Nathan Lile}, \bibinfo{person}{Dakota Mahan}, \bibinfo{person}{Louis Castricato}, \bibinfo{person}{Jan-Philipp Franken}, \bibinfo{person}{Nick Haber}, {and} \bibinfo{person}{Chelsea Finn}.} \bibinfo{year}{2025}\natexlab{}.
\newblock \bibinfo{title}{Towards System 2 Reasoning in LLMs: Learning How to Think With Meta Chain-of-Thought}.
\newblock
\showeprint[arxiv]{2501.04682}~[cs.AI]
\urldef\tempurl%
\url{https://arxiv.org/abs/2501.04682}
\showURL{%
\tempurl}


\bibitem[Xu et~al\mbox{.}(2024)]%
        {thingscringeothersiterative.corr24}
\bibfield{author}{\bibinfo{person}{Jing Xu}, \bibinfo{person}{Andrew Lee}, \bibinfo{person}{Sainbayar Sukhbaatar}, {and} \bibinfo{person}{Jason Weston}.} \bibinfo{year}{2024}\natexlab{}.
\newblock \bibinfo{title}{Some things are more CRINGE than others: Iterative Preference Optimization with the Pairwise Cringe Loss}.
\newblock
\showeprint[arxiv]{2312.16682}~[cs.CL]
\urldef\tempurl%
\url{https://arxiv.org/abs/2312.16682}
\showURL{%
\tempurl}


\bibitem[Yao et~al\mbox{.}(2024)]%
        {cacheblend}
\bibfield{author}{\bibinfo{person}{Jiayi Yao}, \bibinfo{person}{Hanchen Li}, \bibinfo{person}{Yuhan Liu}, \bibinfo{person}{Siddhant Ray}, \bibinfo{person}{Yihua Cheng}, \bibinfo{person}{Qizheng Zhang}, \bibinfo{person}{Kuntai Du}, \bibinfo{person}{Shan Lu}, {and} \bibinfo{person}{Junchen Jiang}.} \bibinfo{year}{2024}\natexlab{}.
\newblock \showarticletitle{CacheBlend: Fast Large Language Model Serving with Cached Knowledge Fusion}.
\newblock \bibinfo{journal}{\emph{arXiv preprint arXiv:2405.16444}} (\bibinfo{year}{2024}).
\newblock


\bibitem[Ye et~al\mbox{.}(2025)]%
        {flashinfer.corr25}
\bibfield{author}{\bibinfo{person}{Zihao Ye}, \bibinfo{person}{Lequn Chen}, \bibinfo{person}{Ruihang Lai}, \bibinfo{person}{Wuwei Lin}, \bibinfo{person}{Yineng Zhang}, \bibinfo{person}{Stephanie Wang}, \bibinfo{person}{Tianqi Chen}, \bibinfo{person}{Baris Kasikci}, \bibinfo{person}{Vinod Grover}, \bibinfo{person}{Arvind Krishnamurthy}, {and} \bibinfo{person}{Luis Ceze}.} \bibinfo{year}{2025}\natexlab{}.
\newblock \showarticletitle{FlashInfer: Efficient and Customizable Attention Engine for LLM Inference Serving}.
\newblock \bibinfo{journal}{\emph{arXiv preprint arXiv:2501.01005}} (\bibinfo{year}{2025}).
\newblock
\urldef\tempurl%
\url{https://arxiv.org/abs/2501.01005}
\showURL{%
\tempurl}


\bibitem[Yu et~al\mbox{.}(2022)]%
        {orca}
\bibfield{author}{\bibinfo{person}{Gyeong-In Yu}, \bibinfo{person}{Joo~Seong Jeong}, \bibinfo{person}{Geon-Woo Kim}, \bibinfo{person}{Soojeong Kim}, {and} \bibinfo{person}{Byung-Gon Chun}.} \bibinfo{year}{2022}\natexlab{}.
\newblock \showarticletitle{Orca: A distributed serving system for $\{$Transformer-Based$\}$ generative models}. In \bibinfo{booktitle}{\emph{{16th USENIX Symposium on Operating Systems Design and Implementation (OSDI 22)}}}. \bibinfo{pages}{521--538}.
\newblock


\bibitem[Zan et~al\mbox{.}(2022)]%
        {cert.corr22}
\bibfield{author}{\bibinfo{person}{Daoguang Zan}, \bibinfo{person}{Bei Chen}, \bibinfo{person}{Dejian Yang}, \bibinfo{person}{Zeqi Lin}, \bibinfo{person}{Minsu Kim}, \bibinfo{person}{Bei Guan}, \bibinfo{person}{Yongji Wang}, \bibinfo{person}{Weizhu Chen}, {and} \bibinfo{person}{Jian-Guang Lou}.} \bibinfo{year}{2022}\natexlab{}.
\newblock \bibinfo{title}{CERT: Continual Pre-Training on Sketches for Library-Oriented Code Generation}.
\newblock
\showeprint[arxiv]{2206.06888}~[cs.SE]
\urldef\tempurl%
\url{https://arxiv.org/abs/2206.06888}
\showURL{%
\tempurl}


\bibitem[Zhang et~al\mbox{.}(2023a)]%
        {copf.corr23}
\bibfield{author}{\bibinfo{person}{Han Zhang}, \bibinfo{person}{Lin Gui}, \bibinfo{person}{Yuanzhao Zhai}, \bibinfo{person}{Hui Wang}, \bibinfo{person}{Yu Lei}, {and} \bibinfo{person}{Ruifeng Xu}.} \bibinfo{year}{2023}\natexlab{a}.
\newblock \showarticletitle{Copf: Continual learning human preference through optimal policy fitting}.
\newblock \bibinfo{journal}{\emph{CoRR}} (\bibinfo{year}{2023}).
\newblock


\bibitem[Zhang et~al\mbox{.}(2023b)]%
        {reformulating.corr23}
\bibfield{author}{\bibinfo{person}{Yating Zhang}, \bibinfo{person}{Yexiang Wang}, \bibinfo{person}{Fei Cheng}, \bibinfo{person}{Sadao Kurohashi}, {et~al\mbox{.}}} \bibinfo{year}{2023}\natexlab{b}.
\newblock \showarticletitle{Reformulating Domain Adaptation of Large Language Models as Adapt-Retrieve-Revise: A Case Study on Chinese Legal Domain}.
\newblock \bibinfo{journal}{\emph{arXiv preprint arXiv:2310.03328}} (\bibinfo{year}{2023}).
\newblock


\bibitem[Zhao et~al\mbox{.}(2024)]%
        {atom.corr24}
\bibfield{author}{\bibinfo{person}{Yilong Zhao}, \bibinfo{person}{Chien-Yu Lin}, \bibinfo{person}{Kan Zhu}, \bibinfo{person}{Zihao Ye}, \bibinfo{person}{Lequn Chen}, \bibinfo{person}{Size Zheng}, \bibinfo{person}{Luis Ceze}, \bibinfo{person}{Arvind Krishnamurthy}, \bibinfo{person}{Tianqi Chen}, {and} \bibinfo{person}{Baris Kasikci}.} \bibinfo{year}{2024}\natexlab{}.
\newblock \showarticletitle{Atom: Low-bit quantization for efficient and accurate llm serving}.
\newblock \bibinfo{journal}{\emph{Proceedings of Machine Learning and Systems}}  \bibinfo{volume}{6} (\bibinfo{year}{2024}), \bibinfo{pages}{196--209}.
\newblock


\bibitem[Zheng et~al\mbox{.}(2024)]%
        {sglang}
\bibfield{author}{\bibinfo{person}{Lianmin Zheng}, \bibinfo{person}{Liangsheng Yin}, \bibinfo{person}{Zhiqiang Xie}, \bibinfo{person}{Chuyue Sun}, \bibinfo{person}{Jeff Huang}, \bibinfo{person}{Cody~Hao Yu}, \bibinfo{person}{Shiyi Cao}, \bibinfo{person}{Christos Kozyrakis}, \bibinfo{person}{Ion Stoica}, \bibinfo{person}{Joseph~E Gonzalez}, {et~al\mbox{.}}} \bibinfo{year}{2024}\natexlab{}.
\newblock \showarticletitle{Sglang: Efficient execution of structured language model programs}.
\newblock \bibinfo{journal}{\emph{arXiv preprint arXiv:2312.07104}} (\bibinfo{year}{2024}).
\newblock


\bibitem[Zhong et~al\mbox{.}(2024a)]%
        {distserve}
\bibfield{author}{\bibinfo{person}{Yinmin Zhong}, \bibinfo{person}{Shengyu Liu}, \bibinfo{person}{Junda Chen}, \bibinfo{person}{Jianbo Hu}, \bibinfo{person}{Yibo Zhu}, \bibinfo{person}{Xuanzhe Liu}, \bibinfo{person}{Xin Jin}, {and} \bibinfo{person}{Hao Zhang}.} \bibinfo{year}{2024}\natexlab{a}.
\newblock \bibinfo{title}{DistServe: Disaggregating Prefill and Decoding for Goodput-optimized Large Language Model Serving}.
\newblock
\showeprint[arxiv]{2401.09670}~[cs.DC]


\bibitem[Zhong et~al\mbox{.}(2024b)]%
        {zhong2024rlhfuseefficientrlhftraining}
\bibfield{author}{\bibinfo{person}{Yinmin Zhong}, \bibinfo{person}{Zili Zhang}, \bibinfo{person}{Bingyang Wu}, \bibinfo{person}{Shengyu Liu}, \bibinfo{person}{Yukun Chen}, \bibinfo{person}{Changyi Wan}, \bibinfo{person}{Hanpeng Hu}, \bibinfo{person}{Lei Xia}, \bibinfo{person}{Ranchen Ming}, \bibinfo{person}{Yibo Zhu}, {and} \bibinfo{person}{Xin Jin}.} \bibinfo{year}{2024}\natexlab{b}.
\newblock \bibinfo{title}{RLHFuse: Efficient RLHF Training for Large Language Models with Inter- and Intra-Stage Fusion}.
\newblock
\showeprint[arxiv]{2409.13221}~[cs.LG]
\urldef\tempurl%
\url{https://arxiv.org/abs/2409.13221}
\showURL{%
\tempurl}


\bibitem[Ziegler et~al\mbox{.}(2020)]%
        {finetuninglanguagemodelshuman.corr20}
\bibfield{author}{\bibinfo{person}{Daniel~M. Ziegler}, \bibinfo{person}{Nisan Stiennon}, \bibinfo{person}{Jeffrey Wu}, \bibinfo{person}{Tom~B. Brown}, \bibinfo{person}{Alec Radford}, \bibinfo{person}{Dario Amodei}, \bibinfo{person}{Paul Christiano}, {and} \bibinfo{person}{Geoffrey Irving}.} \bibinfo{year}{2020}\natexlab{}.
\newblock \bibinfo{title}{Fine-Tuning Language Models from Human Preferences}.
\newblock
\showeprint[arxiv]{1909.08593}~[cs.CL]
\urldef\tempurl%
\url{https://arxiv.org/abs/1909.08593}
\showURL{%
\tempurl}


\end{thebibliography}

\end{document}